\documentclass[11pt,letterpaper,teaser]{planstyle}
\usepackage[T1]{fontenc}
\usepackage{microtype}           % Microtypographic refinements
\usepackage{nicefrac}            % Compact symbols for 1/2, etc.
\usepackage{xspace}              % Smart spacing after macros
\usepackage{amsmath, amssymb, amsfonts, amsthm, mathtools, mathrsfs}
\usepackage{dsfont}              % Double-stroke font
\usepackage{fdsymbol}            % Additional math symbols

\usepackage{booktabs}            % Professional-quality tables
\usepackage{nicematrix}         % Even better sometimes
\usepackage{multirow}
\usepackage{makecell}
\usepackage{bigstrut}
\usepackage{enumitem}            % Customized lists
\setitemize{label=\textbullet, leftmargin=*, nolistsep}
\usepackage{graphicx}
\usepackage{adjustbox}           % Flexible box for figures/tables
\usepackage{float}               % Precise float placement
\usepackage{wrapfig}             % Wrapping figures
\usepackage{subcaption}          % Recommended over subfig
\expandafter\def\csname ver@subfig.sty\endcsname{}  % Avoid conflict with subcaption
\usepackage[dvipsnames]{xcolor}    % Extended color names
\usepackage{color}
\usepackage{fontawesome5}        % Icons
\usepackage{pifont}              % Dingbats
\usepackage{hyperref}            % Hyperlinks
\hypersetup{colorlinks=true, citecolor=LightBlue}
\usepackage[all]{hypcap}         % Fixes links to floats
\usepackage{cleveref}            % Smart cross-referencing
\usepackage{url}                 % For typesetting URLs
\usepackage{algorithm}
\usepackage{algorithmicx}
\usepackage{algpseudocode}
\usepackage[normalem]{ulem}      % Underline, strikeout, etc.
\usepackage{lipsum}              % Dummy text
\usepackage{rotating}            % Rotated objects
\usepackage{stackengine}         % Stack text and symbols
\usepackage{caption}             % Caption customization
\usepackage{listings}            % Code listings
\usepackage{soul}                % Undeline (works with break lines)
\usepackage{gradient-text}       % Gradient colored text
\usepackage{pgfplots}            % latex plots
\pgfplotsset{compat=newest}
\usepackage{pgfplotstable}       % for external data tables
\usepackage{tikz}                % Drawing
\usepackage{svg}                 % Include SVG graphics

\usepackage[comma,numbers,sort,compress]{natbib}
\definecolor{demphcolor}{RGB}{125,125,125}    

\hypersetup{
    colorlinks = true,
    citecolor = {YaleBlue},
}
\tcbuselibrary{breakable, skins}

\newcommand{\ie}{\textit{i.e.},\xspace}      % i.e.
\newcommand{\eg}{\textit{e.g.},\xspace}      % e.g.
\crefname{equation}{Eq.}{Eqs.}
\crefformat{section}{\S#2#1#3}
\crefformat{subsection}{\S#2#1#3}
\crefformat{subsubsection}{\S#2#1#3}
\crefrangeformat{section}{\S\S#3#1#4 to~#5#2#6}
\crefmultiformat{section}{\S\S#2#1#3}{ and~#2#1#3}{, #2#1#3}{ and~#2#1#3}
\newcommand{\cmark}{\textcolor{ForestGreen}{\ding{51}}}  % ✓
\newcommand{\xmark}{\textcolor{red}{\ding{55}}}     % ✗

\newcommand{\bm}{\mathbf{m}}
\def\vx{{\bm{x}}}

\def\vz{{\bm{z}}}

\newcommand{\Enc}{\operatorname{Enc}} 
\usepackage{tcolorbox}
\tcbuselibrary{listings,breakable}
\usepackage{cuted} % for strip environment
\usepackage{dblfloatfix}
\usepackage{hyperref}
\usepackage{url}
\usepackage{wrapfig}
\usepackage{tcolorbox}

\definecolor{fire0}{HTML}{FFF2B2}
\definecolor{fire1}{HTML}{C1121F}
\definecolor{fire2}{HTML}{FFB347}
\definecolor{fire3}{HTML}{FF8A3D}
\definecolor{fire4}{HTML}{FF4D00}
\definecolor{fire5}{HTML}{FF4A3A}
\definecolor{fire6}{HTML}{CE0A18}
\definecolor{UCSDBlue}{HTML}{182B49}
\definecolor{pyraAmber}{HTML}{FFB347}
\definecolor{pyraOrange}{HTML}{FF6B3D}
\definecolor{pyraRed}{HTML}{E63946}

\definecolor{EgoBlue}{HTML}{4C65A4}
\definecolor{EgoAccent}{HTML}{CC4B37}
\definecolor{EgoAccentEnd}{HTML}{913B97}
\definecolor{EgoPeriwinkle}{HTML}{8B8FCA}
\definecolor{EgoLavender}{HTML}{B7AED9}
\definecolor{EgoPink}{HTML}{E8C1CC}
\definecolor{EgoApricot}{HTML}{FFD7A8}
\definecolor{EgoPeach}{HTML}{F3A683}
\definecolor{EgoTerracotta}{HTML}{B9786F}
\colorlet{TableHeader}{EgoBlue!18}
\colorlet{TableGroup}{EgoLavender!25}
\colorlet{TableHighlight}{EgoApricot!45}
\newcommand{\best}[1]{\textbf{#1}}
\definecolor{AppendixColor}{HTML}{4C65A4}

\theoremstyle{plain}

\theoremstyle{definition}

\theoremstyle{remark}

\usepackage{enumitem}
\setlist[itemize]{leftmargin=*}

\newcommand{\tabicon}[1]{\raisebox{-0.9ex}{\includegraphics[height=3ex]{assets/icon/#1}}}

\newtcolorbox{planbox}[2][EgoBlue]{
  enhanced,breakable,lines before break=6,
  title after break={#2 (continued)},colback=#1!5!white,colframe=#1!75!black,
  colbacktitle=#1!20!white,coltitle=#1!35!black,
  title={#2},fonttitle=\bfseries,boxrule=0.6pt,arc=1mm,
  left=9pt,right=9pt,top=7pt,bottom=7pt,
  before skip=10pt,after skip=10pt}
\newcommand{\promptheading}[1]{\textcolor{EgoBlue!75!black}{\textbf{#1}}}

\newcommand{\modelname}{\textbf{EgoForge\xspace}}
\newcommand{\modelnamenc}{EgoForge\xspace}
\newcommand{\datasetname}{\textbf{X-Ego\xspace}}
\newcommand{\datasetnamenc}{{X-Ego\xspace}}
\DeclareRobustCommand{\graft}{\mbox{GRAFT}\xspace}
\DeclareRobustCommand{\graftfull}{Goal-conditioned Rollout Alignment with negative-aware Fine-Tuning}
\title{\vspace{-0.5cm}
EgoForge: Goal-Directed Egocentric World Simulator} 
\author{
\vspace{-0.5cm}
    \textbf{Yifan Shen\textsuperscript{\textcolor{IllinoisOrange}{1}}, Jiateng Liu\textsuperscript{\textcolor{IllinoisOrange}{1}}, Xinzhuo Li\textsuperscript{\textcolor{IllinoisOrange}{1}}, Yuanzhe Liu\textsuperscript{\textcolor{IllinoisOrange}{1}}, Bingxuan Li\textsuperscript{\textcolor{IllinoisOrange}{\textcolor{IllinoisOrange}{1}}}, Houze Yang\textsuperscript{\textcolor{IllinoisOrange}{1}}, Wenqi Jia\textsuperscript{\textcolor{IllinoisOrange}{1}}} \\
    \textbf{Yijiang Li\textsuperscript{\textcolor{UCSDBlue}{2}}, Tianjiao Yu\textsuperscript{\textcolor{IllinoisOrange}{1}}, James Matthew Rehg\textsuperscript{\textcolor{IllinoisOrange}{1}}, Xu Cao\textsuperscript{\textcolor{IllinoisOrange}{1},$\dagger$}, Ismini Lourentzou\textsuperscript{\textcolor{IllinoisOrange}{1},$\dagger$}}  
    }

\affil{\textcolor{IllinoisOrange}{\textsuperscript{1} University of Illinois Urbana-Champaign}, \textcolor{UCSDBlue}{\textsuperscript{2}University of California San Diego}
\vspace{-0.5cm}
}

\correspondingauthor{
  \textsuperscript{$\star$}Equal Contribution \\
  \textsuperscript{$\dagger$}Equal Corresponding Authors
}

\begin{document}

\setabstractlogo[9mm]{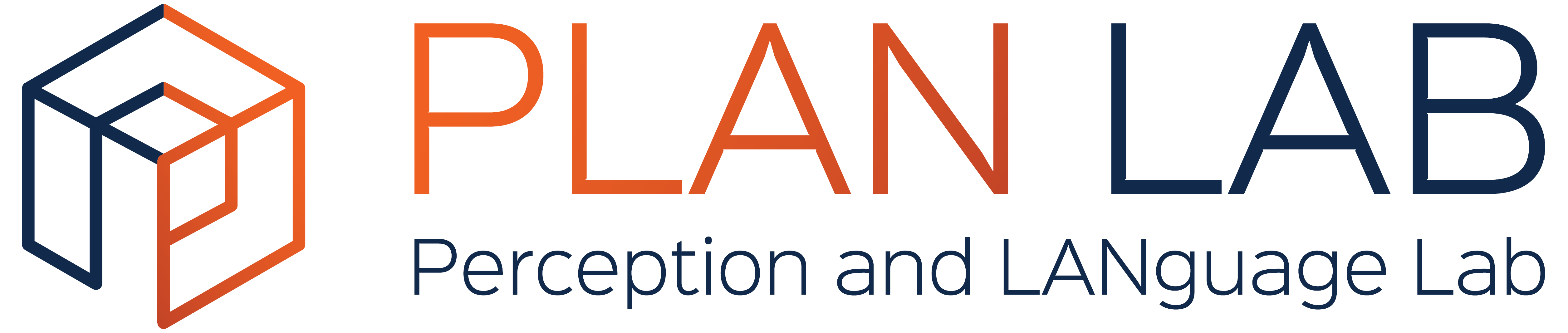} 

\begin{abstract}
Generative world models have shown promise for simulating dynamic environments, yet egocentric video remains challenging due to rapid viewpoint changes, frequent hand–object interactions, and goal-directed procedures whose evolution depends on latent human intent. Existing approaches either focus on hand-centric instructional synthesis with limited scene evolution, perform static view translation without modeling action dynamics, or rely on dense supervision, such as camera trajectories, long video prefixes, and synchronized multi-camera capture. In this work, we introduce \modelname{}, an egocentric goal-directed world simulator that generates coherent, first-person video rollouts from minimal static inputs: a single egocentric image, a high-level instruction, and an optional auxiliary exocentric view.
To improve intent alignment and temporal coherence, we introduce \graft{}, a trajectory-level diffusion refinement method that uses positive and negative rollout distributions, derived from goal, temporal, scene-consistency, and perceptual rewards, to steer the diffusion velocity field toward coherent, goal-complete egocentric simulations.
Extensive experiments show \modelnamenc{} achieves consistent gains in semantic alignment, geometric stability, and motion fidelity over strong baselines, and performs robustly in real-world smart-glasses experiments.

\url{https://plan-lab.github.io/egoforge}
\vspace{-2mm}
\end{abstract}

\renewcommand{\insertteaserfigure}{
    \vspace{-1cm}
      \begin{center}
    \centering
   \includegraphics[width=0.99\textwidth]{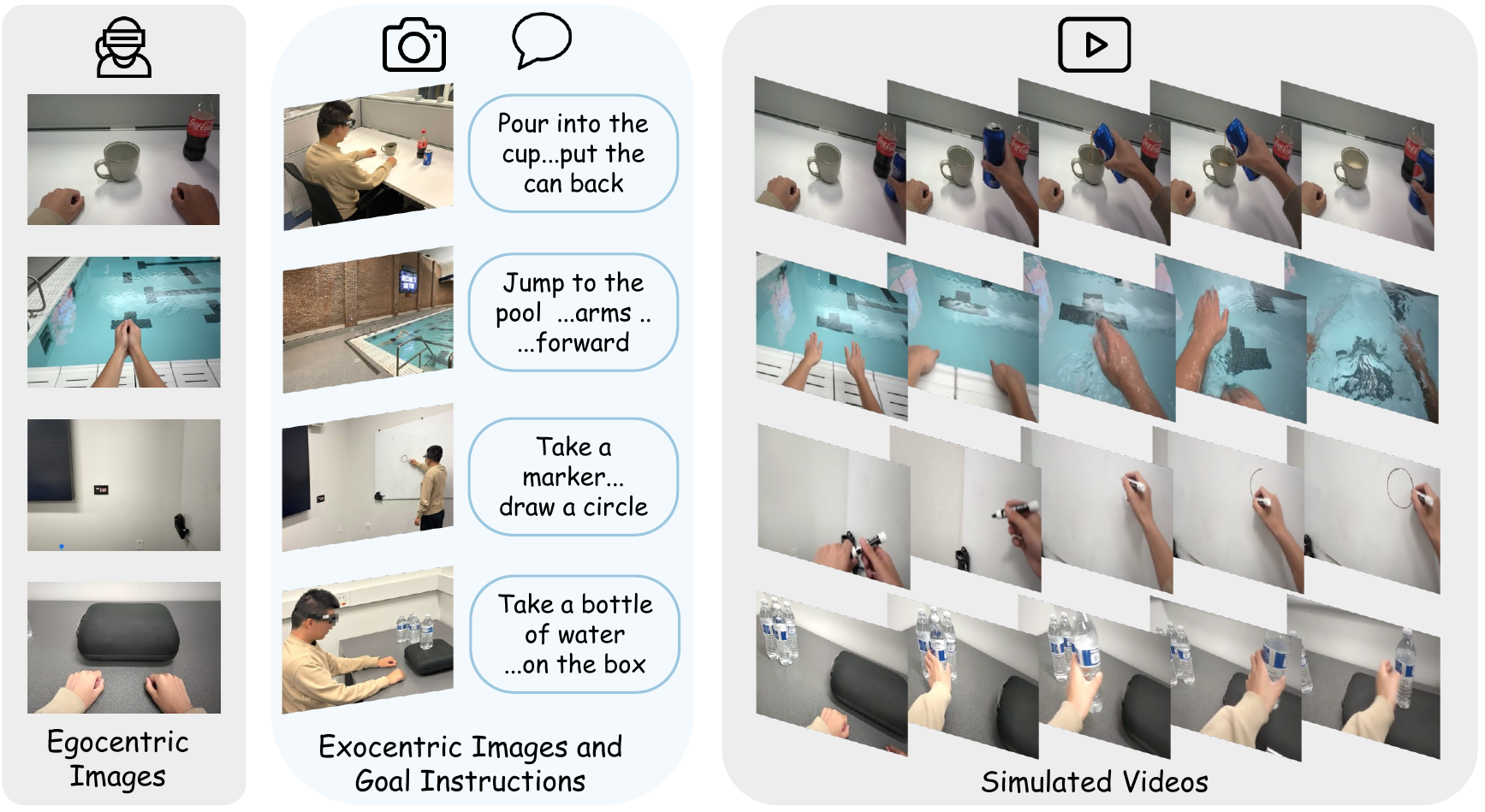}
    %\vspace{-0.2cm}
    \captionof{figure}{\textbf{Egocentric video rollouts produced by \modelname{} in real-world smart-glasses experiments.} Given a single smart-glasses egocentric image, a high-level goal instruction, and an auxiliary exocentric view, \modelnamenc{} generates egocentric rollouts that follow user intent and preserve scene structure, without requiring dense supervision, such as camera trajectories, video, or synchronized multi-view capture streams.}
    \label{fig:teaser}
\end{center}
}

\maketitle

% ===================== Main Sections =====================
\section{Introduction}
Generative world models are redefining how AI systems learn, simulate, and reason about dynamic environments~\citep{matsuo2022deep,hu2023gaia,bruce2024genie}. Recent work has made progress in natural scenes generation, such as autonomous driving~\citep{zhou2024simgen,russell2025gaia}, embodied navigation~\citep{bar2025navigation}, and virtual worlds~\citep{ball2025genie3}.
However, despite their visual fidelity and predictive power, these models struggle to model human actions in egocentric vision. 
First-person streams exhibit rapid viewpoint changes, frequent hand–object interactions, and goal-directed behaviors whose evolution depends on latent human intent.
Simulating first-person experiences requires understanding affordances and procedural dependencies underlying human actions, which traditional world models overlook.

These limitations have become increasingly important as the demand for immersive and interactive experiences grows across Extended Reality (XR) platforms, including virtual and augmented environments~\citep{tu2025playerone,zhang2025matrix,yang2025contextagent}. Such applications require human-centric world models capable of generating predictive, controllable, and physically consistent simulations. Egocentric world models could enable next-generation immersive digital experiences. Existing approaches to egocentric generation~\citep{zhang2025egocentricpredictivemodelconditioned,tu2025playerone,zhang2025matrix,agarwal2025cosmos,team2025aether} face three fundamental limitations:
\textbf{(1) Dense supervision requirements.} Existing methods typically require dense motion annotations, calibrated trajectories, video prefixes, or synchronized multi-view recordings, which are costly to obtain at scale and cannot be assumed to be available at inference time in wearable scenarios.
\textbf{(2) Limited goal-directed control.} Most models are conditioned on short textual prompts or low-level actions (\eg keyboard or joint controls), producing generic motion patterns and offering limited control over multi-step behaviors. As a result, they struggle to represent semantic intent such as ``open the fridge and pour milk.''
\textbf{ (3) Weak physical grounding.} Video diffusion models are optimized for visual realism but lack 3D awareness, preventing consistent reasoning about egocentric motion.

To address these challenges, we introduce \textbf{\modelname{}}, an egocentric world simulator 
designed for realistic and controllable first-person video generation.
Unlike existing approaches that rely on dense motion signals, multi-view capture, or video input, \modelnamenc{} generates coherent, goal-directed egocentric rollouts from a single egocentric observation, a high-level instruction, and an optional auxiliary exocentric reference image. 
Built upon a diffusion-transformer backbone,
\modelnamenc{} enforces representational alignment~\citep{yu2025representationalignmentgenerationtraining,wu2025geometryforcingmarryingvideo}  between diffusion latents and 3D geometric features for spatially stable video synthesis.
To improve rollout behavior and better align generation with task intent, we further propose \graft{}, a trajectory-level reward-guided refinement stage that uses positive and negative
rollout distributions, derived from goal, scene-consistency, temporal-causality, and perceptual
rewards, to steer the diffusion velocity field toward coherent, goal-complete egocentric simulations.

To evaluate goal-directed world simulation, we curate \datasetname{}, a benchmark pairing egocentric observations with semantic annotations for grounded video generation.
Comprehensive experiments demonstrate that
\modelname{} achieves large gains in semantic alignment while substantially improving video realism and temporal coherence, with 43\% lower FVD and 51\% lower Flow MSE compared to strong baselines. 
\modelnamenc{} also yields higher structural fidelity (+9.7\% SSIM ↑), lower perceptual error (-35\% LPIPS ↓), and improved reconstruction quality (+17.8\% PSNR ↑). 

The contributions of our work are:\looseness-1

\begin{itemize}[itemsep=0.5ex, parsep=0pt, topsep=-3pt, leftmargin=0.7cm]
    \item We introduce \modelname{}, an egocentric world simulator that generates goal-directed first-person video rollouts from minimal inputs, \ie a single egocentric image, a high-level instruction, and an optional auxiliary exocentric view. To our knowledge, this is the first work to generate goal-directed egocentric rollouts beyond hand-centric instructional motion from minimal static context without pose/trajectory inputs, video prefixes, or synchronized multi-view capture at inference.
    \item To improve intent alignment and temporal coherence, we propose \textbf{\graft{}}, a trajectory-level reward-guided refinement mechanism for video diffusion that fuses goal completion, scene stability, temporal causality, and perceptual fidelity into a unified vector-field update steering sampling toward coherent, goal-consistent rollouts.\looseness-1
    \item We curate the \datasetname{} benchmark, pairing egocentric observations with detailed event flows, hand--object interactions, object-state changes, and auxiliary visual references, enabling systematic evaluation of goal alignment, temporal coherence, and physical consistency.
\end{itemize}
\Needspace{28\baselineskip}

\section{Related Work}
\label{sec:appen_related_works}
\begin{wraptable}{r}{0.5\textwidth}
\vspace{-\intextsep}
\centering
\setlength{\tabcolsep}{3pt}
\renewcommand{\arraystretch}{1.1}
\caption{
\textbf{Representative Related Works.}  I/O icons represent  \tabicon{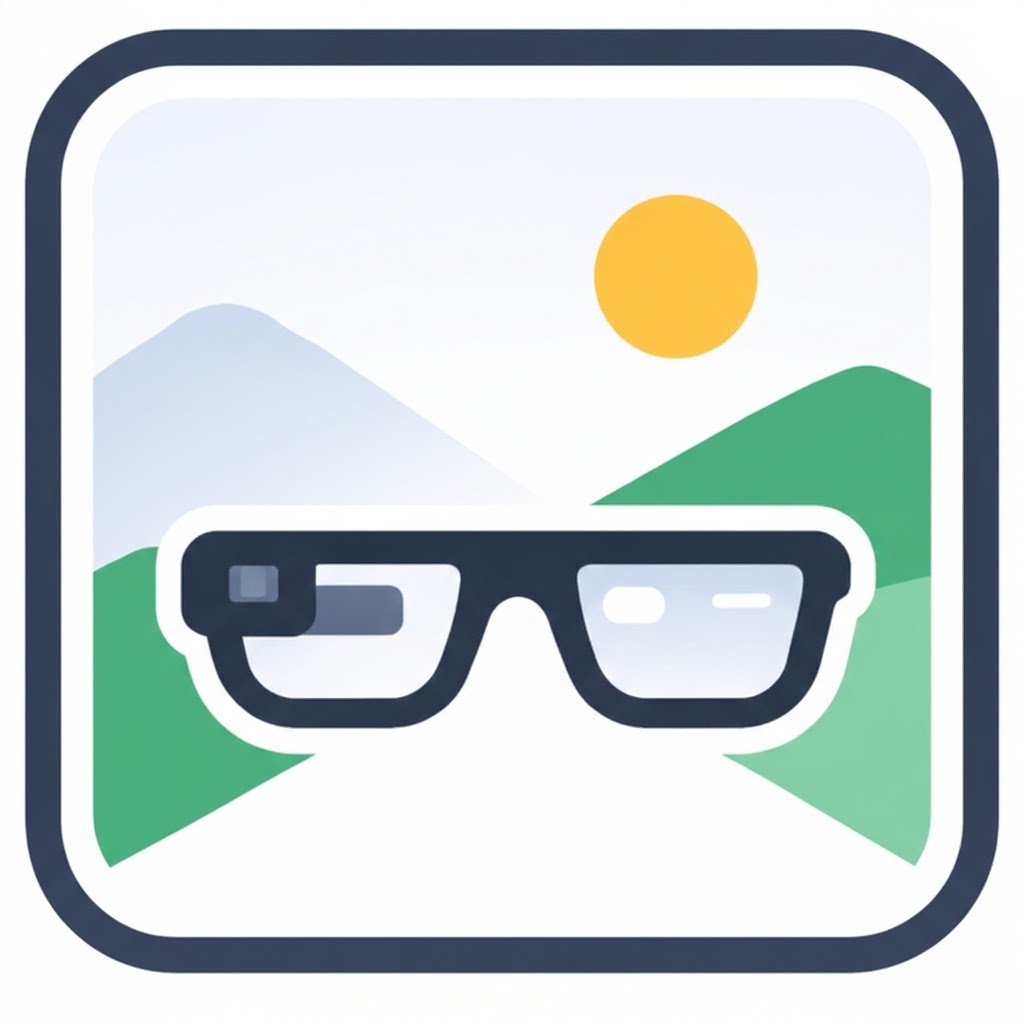} Ego-view, \tabicon{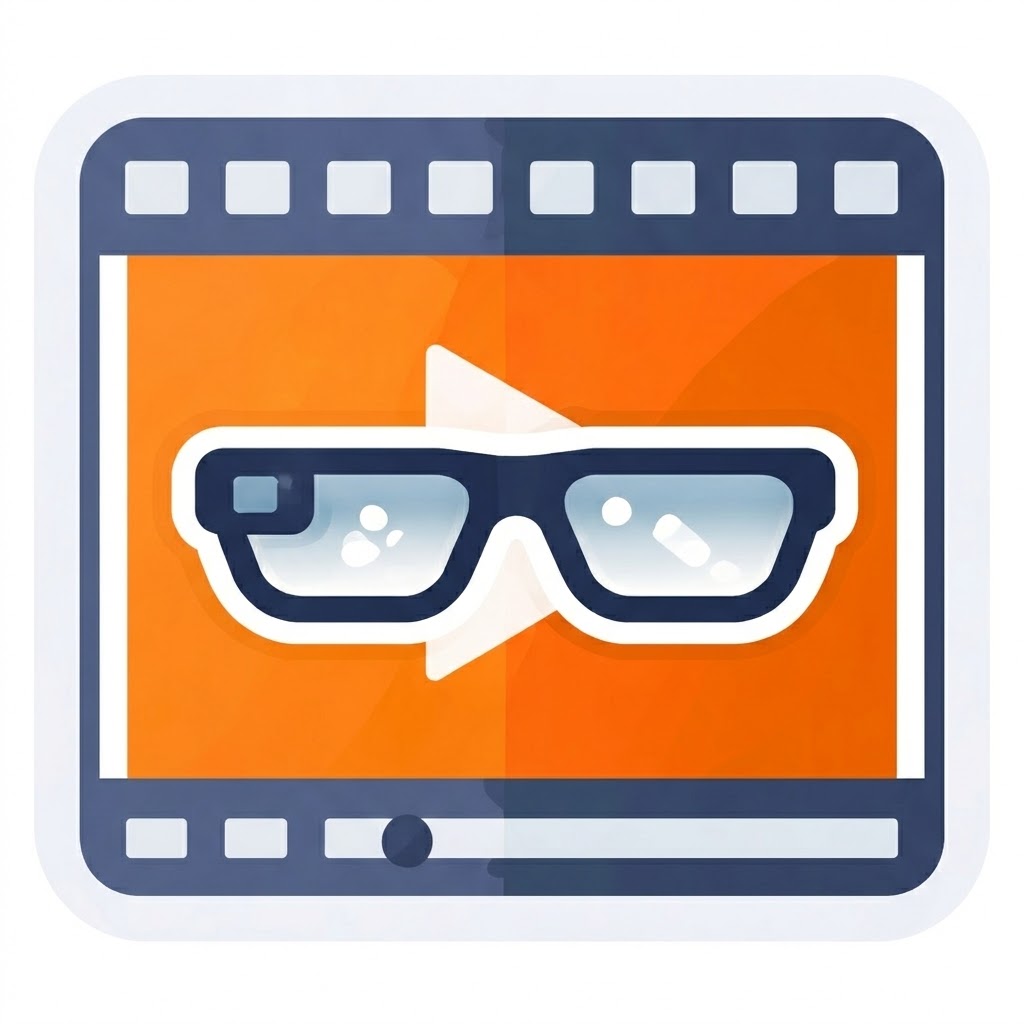} Ego Video, \tabicon{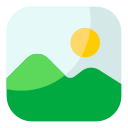} Exo Image, \tabicon{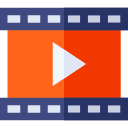} Exo Video stream, \tabicon{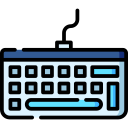} Text/Action prompt, and \tabicon{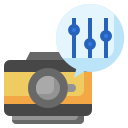} Camera parameters. 
Prior work requires (multi-view) video streams or camera trajectories, while \modelnamenc{} generates from minimal static observations. $^\ddagger$assumes fixed camera poses. $^\star$hand motion synthesis.}
\label{tab:io_comparison}
%\vspace{-0.3cm}
\begin{adjustbox}{width=\linewidth}
\begin{tabular}{lll}
\toprule
\rowcolor{TableHeader}
\textbf{Model} & \textbf{Inputs} & \textbf{Output} \\
\midrule
4Diff~\citep{cheng20244diff} &
\tabicon{Image} + \tabicon{Camera_motion_parameters} &
\tabicon{egoimage} \\

EgoWorld~\citep{park2025egoworld} &
\tabicon{Image} &
\tabicon{egoimage} \\

Exo2Ego-V~\citep{zhang2025exo2ego} &
\tabicon{Video} $\times$ 4 sync &
\tabicon{egovideo} \\

EgoX$^\ddagger$~\citep{kang2025egox} &
\tabicon{Video}&
\tabicon{egovideo} \\

Handi$^\star$~\citep{li2024handi} &
\tabicon{egoimage} + \tabicon{Keyboard} &
\tabicon{egovideo} \\

EgoDreamer~\citep{wang2024egovid} &
\tabicon{egoimage} + \tabicon{Keyboard} + \tabicon{Camera_motion_parameters} &
\tabicon{egovideo} \\

\midrule
\rowcolor{TableHighlight}
\textbf{\modelnamenc{} (Ours)} &
\tabicon{egoimage} +  \tabicon{Keyboard} +\tabicon{Image}  &
\tabicon{egovideo} \\
\bottomrule
\end{tabular}
\end{adjustbox}
\vspace{-0.3cm}
\end{wraptable}  
As summarized in Table~\ref{tab:io_comparison}, prior work relies on continuous exocentric streams for synchronized view generation~\citep{xu2025egoexo,zhang2025exo2ego,kang2025egox}, performs static view translation~\citep{cheng20244diff,park2025egoworld}, or generates videos in hand-centric settings with largely static scenes~\citep{li2024handi}. Broader egocentric generation typically requires explicit motion signals or synchronized multi-view videos~\citep{zhang2025exo2ego,kang2025egox,li2024handi,wang2024egovid}. In contrast, \modelnamenc{} predicts controllable, physically consistent egocentric videos from a single first-person image and instruction, without dense motion supervision.  

\noindent \textbf{Egocentric Vision.} Egocentric vision research has rapidly advanced with improvements in both dataset quality and model scaling. Foundational benchmarks such as EPIC-KITCHENS~\citep{damen2022rescaling, damen2020epic, damen2018scaling}, Ego4D~\citep{grauman2022ego4d}, and EgoExo4D~\citep{grauman2024ego} have enabled large-scale analysis of daily tasks and multi-view human activity. On the modeling side, prior work focuses on recognizing actions, gaze, attention, and human-object interactions from egocentric views~\citep{huang2018predicting, kazakos2019epic, wang2021interactive, liu2022joint, ragusa2023stillfast}, or on estimating human body pose~\citep{li2023ego, luo2021dynamics, wang2023scene}. 
There is also increasing attention to vision–language models for learning multimodal representations from egocentric data~\citep{lin2022egocentric,pramanick2023egovlpv2,ashutosh2023hiervl,cao2024visual,shen2025fine} and forecasting plausible hand motions~\citep{jia2022generative}, hand-object interactions~\citep{ye2023affordance}, and gaze~\citep{zhang2017deep}. 
Methods for cross-view exocentric-to-egocentric translation and joint egocentric video-motion synthesis depend on explicit motion supervision, such as camera trajectories or synchronized multi-camera exocentric recordings~\citep{cheng20244diff,liu2024exocentric,xiu2025egotwin,park2025egoworld}. However, such data are difficult to acquire and still fail to capture high-level, goal-driven human intent, limiting controllability and semantic diversity in generated egocentric videos. In contrast, {\modelnamenc{}} generates egocentric videos without requiring motion or view supervision.\looseness-1

\noindent \textbf{Video Generation.} Advances in latent diffusion~\citep{rombach2022high, podell2024sdxl, chen2024pixartalpha} and score-based generative modeling ~\citep{song2019generative, song2021scorebased, ho2020denoising} established the principles of iterative denoising and latent-space synthesis that now underpin most video generators. Building on these foundations, transformer-based video diffusion architectures enable greater temporal coherence, longer video duration, and improved realism~\citep{openai2024sora, peebles2023scalable, kong2024hunyuanvideo, xie2024sana}.
Large-scale frameworks such as Stable Video Diffusion~\citep{blattmann2023stable}, VideoCrafter~\citep{chen2023videocrafter1, chen2024videocrafter2}, and Open-Sora~\citep{lin2024open} demonstrate strong cross-domain synthesis through efficient latent modeling and scaling. While these models excel at general-purpose video generation, they lack representations of agent intent, viewpoint dynamics, and causal continuity required for egocentric simulation. 

\noindent \textbf{World Models.} World models are predictive models trained to simulate the environment dynamics, serving either as tools for agent policy optimization~\citep{hafner2019learning,gorodetskiy2024model,li2024think2drive} or as high-fidelity world simulators and game engines~\citep{alonso2024diffusion, valevski2025diffusion, bruce2024genie, che2025gamegenx,liu2026palm,yu2025core3d}. 
Recent efforts such as Matrix-Game~\citep{zhang2025matrix}, Cosmos~\citep{agarwal2025cosmos}, and Aether~\citep{team2025aether} generate unlimited and interactive video streams of 3D worlds, but lack explicit understanding of user intent and task context. 
Reward-based post-training of diffusion models has been explored for images through likelihood-free or forward-process objectives, as in DiffusionNFT~\citep{zheng2025diffusionnftonlinediffusionreinforcement}, but these methods score individual samples and do not address goal completion over temporally extended rollouts.
In contrast, {\modelnamenc{}} generates an egocentric video trajectory that fulfills a specified goal through our proposed \graft{} trajectory-level reward-guided refinement that optimizes goal completion and temporal coherence.

\section{\modelnamenc{} Method}
\label{sec:method}
\modelnamenc{} is an egocentric world simulator designed to generate goal-directed first-person video rollouts that simulate how a scene evolves when a user performs a specified task. Given an initial egocentric frame or short clip $\vx_{1:k}$, the objective is to synthesize a plausible future sequence $\vx_{k+1:T}$. 
Formally, \modelnamenc{} models\looseness-1 
\begin{equation}
\setlength{\abovedisplayskip}{2pt}
\setlength{\abovedisplayskip}{2pt}
p_\theta(\vx_{k+1:T} \mid \vx_{1:k}) =
\prod_{i=k+1}^{T} p_\theta(\vx_i \mid \vx_{<i}, \mathcal{C}),
\label{eq:conditional-generation}
\end{equation}
where $\mathcal{C}=\{\vx_{1:k}, y, \vx^{exo}\}$ is the conditioning context, which consists of the embeddings derived from $\vx_{1:k}$, instruction $y$, and the exocentric reference $\vx^{exo}$. Unlike prior approaches, we do not assume access to camera trajectories, pose signals, or synchronized multi-view streams at inference time. The generated rollout should satisfy three key properties, such as goal alignment (\ie the sequence reflects the intended instruction), temporal coherence (\ie motion evolves smoothly across frames), and physical consistency (\ie scene geometry and interactions remain stable).

\begin{figure*}[!t]
    \centering
    \includegraphics[width=0.99\textwidth]{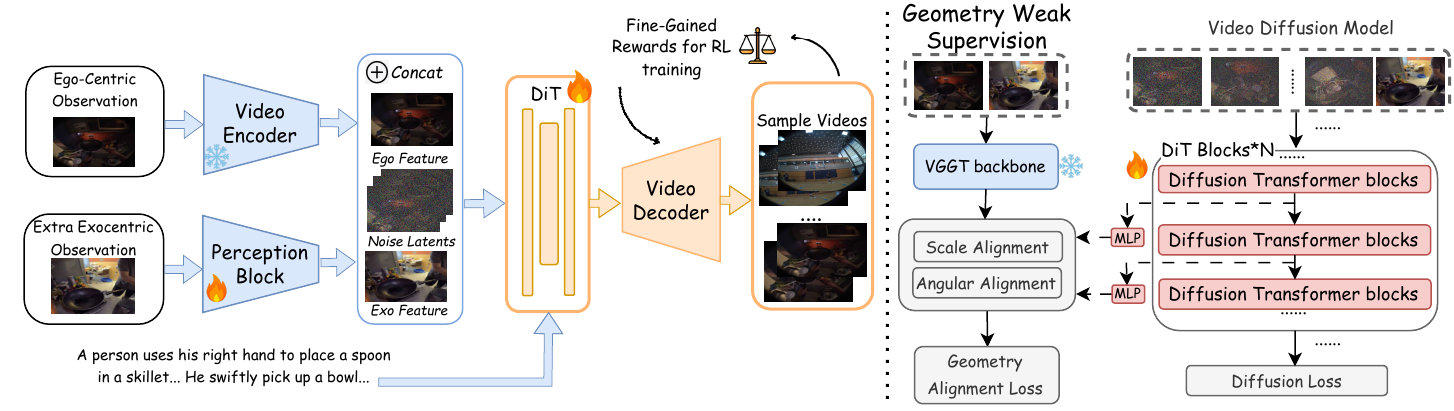}
    \caption{\textbf{\modelnamenc{} Overview}: 
    Given a single egocentric observation, a high-level instruction expressing user intent, and an auxiliary exo-view reference, \modelnamenc{} fuses encoded visual features with noisy video latents at each DiT block to guide generation. Geometry alignment weakly supervises intermediate features using angular and scale consistency to encourage spatially stable rollouts.  The resulting rollout videos are further refined via the proposed \graft{} alignment that optimizes goal completion, scene consistency, temporal causality, and perceptual fidelity.} 
    \label{fig:framework}
\end{figure*}

\subsection{Diffusion-Based Egocentric Generator}\label{sec:model}
Figure \ref{fig:framework} illustrates the overall architecture of \modelnamenc{}, where video generation is performed in the latent space of a pretrained video autoencoder.
Let $\vz_0\!=\!\Enc(\vx_{k+1:T})$ be the latent trajectory. 
We adopt the rectified-flow form of flow matching~\citep{lipman2022flow,liu2022flow}, consistent with the flow-matching pretraining of the Wan2.2 backbone~\citep{wan2025wan}. We sample a continuous time $t\sim\mathcal{U}(0,1)$ and a noise tensor $\epsilon\sim\mathcal{N}(\mathbf{0},\mathbf{I})$, and form the noisy latent by linear interpolation between data and noise:
\begin{equation}
\setlength{\abovedisplayskip}{3pt}
\setlength{\belowdisplayskip}{3pt}
\vz_t = (1-t)\,\vz_0 + t\,\epsilon,
\qquad
\boldsymbol{v} = \frac{\mathrm{d}\vz_t}{\mathrm{d}t} = \epsilon - \vz_0,
\label{eq:forward}
\end{equation}
where $\boldsymbol{v}$ is the constant velocity of the straight path from $\vz_0$ to $\epsilon$, and $t\!=\!1$ corresponds to pure noise.
A diffusion transformer $\boldsymbol{v}_\theta$ is trained to predict this velocity.
Conditioning on $\mathcal{C}$ is implemented through adaptive normalization and cross-attention layers that incorporate the fused embedding. We concatenate the noisy latent with ego and conditioning context features along the channel dimension $\tilde{\vz}_t = \text{Concat}(\vz_t, \mathbf{f}_{\text{ego}}, \mathbf{f}_{\mathcal{C}})$ and inject the time $t$ through a learned time embedding $\gamma(t)$.
The denoising objective is the velocity-matching loss $\mathcal{L}_{\text{D}}\!=\!
\mathbb{E}_{t,\vz_0,\epsilon}
\left[\|\boldsymbol{v}_\theta(\tilde{\vz}_t, t, \mathcal{C}) - (\epsilon - \vz_0)\|_2^2\right],$
and a rollout is generated by integrating the learned ODE $\mathrm{d}\vz_t/\mathrm{d}t\!=\!\boldsymbol{v}_\theta(\tilde{\vz}_t,t,\mathcal{C})$ from $\vz_1\!\sim\!\mathcal{N}(\mathbf{0},\mathbf{I})$ to $t\!=\!0$ and decoding $\vz_0$.

\noindent \textbf{Geometry Weak Supervision.}
To inject 3D reasoning into the diffusion backbone, we align its intermediate representations with geometry features extracted from a pretrained VGGT model~\citep{wang2025vggtvisualgeometrygrounded}, following REPA~\citep{yu2025representationalignmentgenerationtraining} and Geometry Forcing~\citep{wu2025geometryforcingmarryingvideo}. 
Let $\mathbf{g}_{l} \in \mathbb{R}^{N\times Q\times D_g}$ denote the VGGT features at layer $l$,
where $L$ is the number of selected layers, $N$ and $Q$ index temporal and spatial tokens,
and $D_g$ denotes the feature dimension.
For the diffusion transformer, we extract hidden activations
$\mathbf{h}_l \in \mathbb{R}^{N'\times Q'\times D_h}$ from the corresponding layers.
Because the two backbones operate at different resolutions, we introduce a learnable projection operator
$\Pi_l\!:\!\mathbb{R}^{N'\times Q'\times D_h} \rightarrow \mathbb{R}^{N\times Q\times D_g}$,
implemented via spatiotemporal resampling and channel projection, and obtain projected diffusion features $\mathbf{p}_l = \Pi_l(\mathbf{h}_l).$
To encourage the diffusion features to match the direction of VGGT geometry features, we employ a cosine alignment loss
\begin{equation}
\setlength{\abovedisplayskip}{3pt}
\setlength{\belowdisplayskip}{3pt}
    \mathcal{L}^{\text{ang}}
    =
    -\,\frac{1}{L N Q}
    \sum_{l,n,q}
    \cos\!\big(\mathbf{g}_{l,n,q},\,\mathbf{p}_{l,n,q}\big).
\end{equation}
To constrain the magnitude of geometric features and avoid scale collapse, we introduce a scale alignment loss.
We first normalize the projected features as $\tilde{\mathbf{p}}_{l} = \mathbf{p}_{l} / (\|\mathbf{p}_{l}\|_2 + \varepsilon)$ and obtain geometry predictions
$\hat{\mathbf{g}}_{l} = \rho_{l}(\tilde{\mathbf{p}}_{l})$ via a learned linear head $\rho_{l}$.
The scale alignment loss is then defined as
\begin{equation}
\setlength{\abovedisplayskip}{2pt}
\setlength{\belowdisplayskip}{2pt}
    \mathcal{L}^{\text{sca}}
    =
    \frac{1}{L N Q}
    \sum_{l,n,q}
    \big\|
        \hat{\mathbf{g}}_{l,n,q} - \mathbf{g}_{l,n,q}
    \big\|_2^2.
\end{equation}
The geometry-level coordination objective aggregates both terms across a selected set of layers $\mathcal{L}_{\text{G}}\!=\!\zeta_1 \mathcal{L}^{\text{ang}} + \zeta_2 \mathcal{L}^{\text{sca}}$, where $\zeta_1$ and $\zeta_2$ are coefficients balancing the angular and scale consistency.

\subsection{\graft{}: Goal-conditioned Rollout Alignment} \label{VDNFT}
\begin{figure*}[t]
    \centering
    \includegraphics[width=0.99\textwidth]{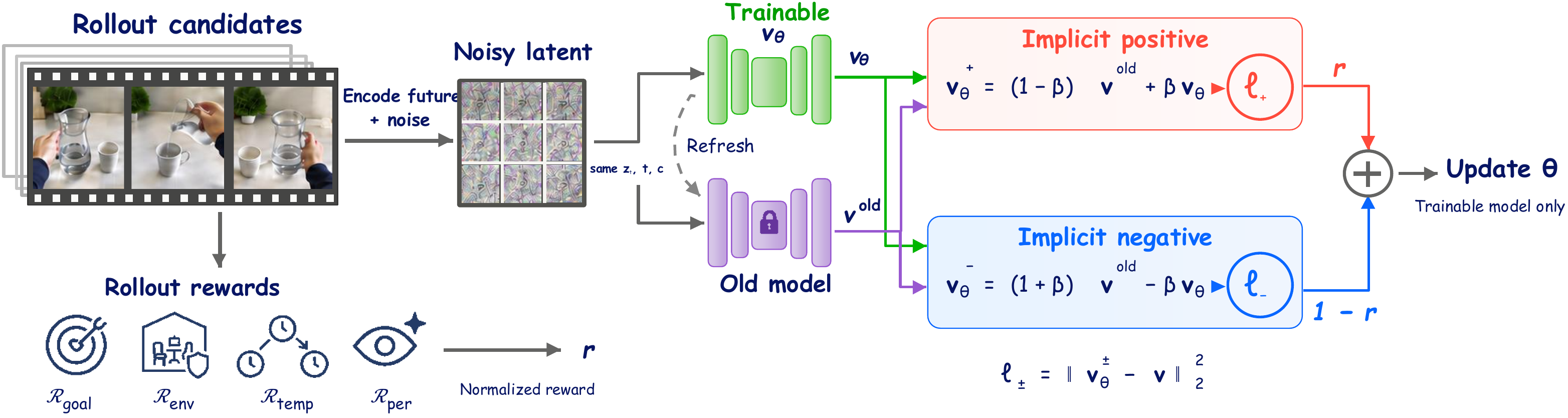}
    \caption{
    \textbf{\graft{} refinement.} Rollout-level rewards yield normalized weights $r$ and $1-r$ for the implicit positive and negative branch losses. Only the trainable model receives gradients, while the old model is refreshed across iterations.}
    \label{fig:graft}
\end{figure*}

Although \modelnamenc{} conditions generation on an egocentric observation, an instruction, and an optional exocentric reference image, these signals do not necessarily contribute equally during denoising. One cue may dominate the sampling trajectory, causing the generated rollout to preserve scene appearance while ignoring the goal, follow the instruction while drifting from the environment, or produce locally plausible motion with incorrect temporal order.
To mitigate this issue, we introduce \graft{} (\graftfull{}), a trajectory-level reward-guided refinement for egocentric video diffusion. \graft{} defines optimality over complete goal-conditioned rollouts rather than individual frames or samples. It scores each rollout with fine-grained rewards for goal completion, scene consistency, temporal causality, and perceptual fidelity. These scores are then converted into per-condition normalized optimality signals that induce positive and negative rollout distributions. The resulting negative-aware update acts directly on the diffusion velocity field.
The positive/negative velocity parameterization and the optimality transform of Eq.~\eqref{eq:optimality} below follow the forward-process formulation of DiffusionNFT~\citep{zheng2025diffusionnftonlinediffusionreinforcement};
the rollout-level optimality, the multi-reward composition of $\mathcal{R}_{\text{total}}$, and the per-condition reward scale $Z_c$ are specific to \graft{}. Figure~\ref{fig:graft} illustrates a refinement iteration of
\graft{}, where rollout-level rewards guide a negative-aware
update of the video generator.

Given the supervised-finetuned policy $\pi^{\mathrm{old}}$, for each condition $c\in\mathcal{C}$ we draw $K$ candidate rollouts $\vx^{(k)}_{1:T}\sim\pi^{\mathrm{old}}(\cdot\mid c)$, $k\!=\!1,\ldots,K$, and assign each rollout a raw reward $R_c^{(k)}\!=\!\mathcal{R}_{\text{total}}(\vx^{(k)}_{1:T},c)$. We estimate the per-condition expected reward with the empirical baseline
$\mu_c\!=\!\frac{1}{K}\sum_{k=1}^{K}R_c^{(k)}\approx\mathbb{E}_{\vx\sim\pi^{\mathrm{old}}(\cdot\mid c)}[\mathcal{R}_{\text{total}}(\vx,c)]$,
and normalize each centered reward into an optimality probability:
\begin{equation}
\setlength{\abovedisplayskip}{3pt}
\setlength{\belowdisplayskip}{3pt}
\scalebox{1}{$
\tilde{\mathcal{R}}^{(k)}_{\text{total}}
=
\tfrac{1}{2}\!\left[
  1 + \operatorname{clip}\!\left(
    \frac{\mathcal{R}^{(k)}_{\text{total}}(\vx^{(k)}_{1:T}, c) - \mu_c}{Z_c},
    -1, 1
  \right)
\right],
$}
\label{eq:optimality}
\end{equation}
where $Z_c > 0$ is a per-condition reward scale, the standard deviation of the $K$ rewards of condition $c$ (Appendix~\ref{sec:implement}), so that
$\tilde{\mathcal{R}}^{(k)}_{\text{total}} \in [0,1]$.
For notational convenience, we denote the normalized optimality as
$r(\vx^{(k)}, c) := \tilde{\mathcal{R}}^{(k)}_{\text{total}}$, so that
$r(\vx^{(k)}, c) \in [0,1]$.
Let $p_{\pi^{\mathrm{old}}}(o=1 \mid c) := \mathbb{E}_{\vx \sim \pi^{\mathrm{old}}(\cdot \mid c)}[r(\vx,c)]$ denote the expected per-condition optimality mass, which normalizes the reweighted positive and negative posteriors:
\begin{equation}
\setlength{\abovedisplayskip}{3pt}
\setlength{\belowdisplayskip}{3pt}
\begin{aligned}
\pi^{+}(\vx \mid c)
=
\frac{r(\vx,c)}{p_{\pi^{\text{old}}}(o=1 \mid c)+\varepsilon}
\, \pi^{\text{old}}(\vx \mid c),
\\[4pt]
\pi^{-}(\vx \mid c)
=
\frac{1 - r(\vx,c)}{1 - p_{\pi^{\text{old}}}(o=1 \mid c) + \varepsilon}
\, \pi^{\text{old}}(\vx \mid c),
\end{aligned}
\end{equation}
where $\varepsilon\!>\!0$ is a small constant to avoid division by zero. By construction, these posteriors satisfy $\pi^{+} \succ \pi^{\text {old }} \succ \pi^{-}$ in expected reward.
We define reinforcement guidance as a vector field acting on the intermediate forward states $
\vz_t$, steering the model toward $\pi^{+}$ while repelling it from $\pi^{-}$. 

Let $\boldsymbol{v}^{+}, \boldsymbol{v}^{-}$, and $\boldsymbol{v}^{\text {old }}$ denote the velocity fields of the respective policies, and let $\alpha(\mathbf{z}_t, c)\!=\!\mathbb{E}[r(\mathbf{x}, c) \mid \mathbf{z}_t, c]$ denote the conditional optimality at the intermediate state $\vz_t$. The improvement direction is
$\Delta(\vz_t, c, t)\!=\!\bigl[1-\alpha(\vz_t,c)\bigr]\bigl(\boldsymbol{v}^{\text{old}}-\boldsymbol{v}^{-}\bigr)\!=\!\alpha(\vz_t,c)\bigl(\boldsymbol{v}^{+}-\boldsymbol{v}^{\text{old}}\bigr),
$ yielding the guided target field $\boldsymbol{v}^*\left(\vz_t, c, t\right)\!=\!\boldsymbol{v}^{\text{old}}\left(\vz_t, c, t\right)+\lambda\, \Delta\left(\vz_t, c, t\right)$
with guidance strength $\lambda\!>\!0$. Rather than learning $\Delta$ as a separate model, we optimize the policy directly via a negative-aware flow-matching loss
\begin{equation}
\setlength{\abovedisplayskip}{3pt}
\setlength{\belowdisplayskip}{3pt}
\mathcal{L}(\theta)
=
\operatorname*{\mathbb{E}}_{c,\,\vx\sim\pi^{\mathrm{old}}(\cdot\mid c),\,t,\,\epsilon}\Bigl[
  r(\vx,c)\left\|\boldsymbol{v}_\theta^{+}-\boldsymbol{v}\right\|_2^2
  + \bigl(1-r(\vx,c)\bigr)\left\|\boldsymbol{v}_\theta^{-}-\boldsymbol{v}\right\|_2^2
\Bigr],
\label{eq:nft_loss}
\end{equation}
where $\boldsymbol{v}\!=\!\epsilon-\vz_0$ is the forward-process velocity target of Eq.~\eqref{eq:forward} evaluated on the generated frames of the collected rollout, $\vz_0\!=\!\Enc(\vx_{k+1:T})$, with the observed frames entering only through $\mathcal{C}$, and
$\boldsymbol{v}_\theta^{+}\!=\!(1-\beta) \boldsymbol{v}^{\text{old}} + \beta \boldsymbol{v}_\theta, \quad \boldsymbol{v}_\theta^{-}\!=\!(1+\beta) \boldsymbol{v}^{\text{old}} - \beta \boldsymbol{v}_\theta$ are implicit positive and negative policies expressed through the single network $\boldsymbol{v}_\theta$ around the frozen $\boldsymbol{v}^{\text{old}}$. Only clean rollouts and their rewards enter the loss, so no likelihood of the sampling chain is required. For a rollout with optimality $r$, the minimizer of Eq.~\eqref{eq:nft_loss} satisfies
\begin{equation}
\setlength{\abovedisplayskip}{2pt}
\setlength{\belowdisplayskip}{2pt}
\boldsymbol{v}_{\theta^*}
=
\boldsymbol{v}^{\text{old}}
+
\frac{2\,r(\vx,c)-1}{\beta}
\bigl(\boldsymbol{v}-\boldsymbol{v}^{\text{old}}\bigr),
\label{eq:nft_opt}
\end{equation}
so rollouts with $r\!>\!\tfrac12$ pull the velocity field toward their own trajectories while rollouts with $r\!<\!\tfrac12$ push it away; in expectation over $\pi^{\mathrm{old}}$ the minimizer of Eq.~\eqref{eq:nft_loss} is the guided target field $\boldsymbol{v}^*$ with $\lambda\!=\!2/\beta$, so $\beta$ sets the guidance strength. Appendix \ref{sec:appendix_complexity} provide more information about \graft{}.

\noindent \textbf{Rewards.}
Given a set of rollout candidates $\mathcal{X} = \{ \vx^{(k)}_{1:T} \}_{k=1}^K$, we evaluate each generated
video trajectory $\vx^{(k)}_{1:T}$ of length $T$ with a set of scalar rewards that measure goal completion, environment preservation, temporal causality, and perceptual fidelity:

\begin{itemize}[itemsep=0ex, parsep=1pt, topsep=0pt, leftmargin=2em]
    \item[\faGlasses] \textbf{Goal Completion} \((\mathcal{R}_{\text{goal}})\) evaluates task completion, \ie whether the trajectory successfully achieves the task outcome, measured by the similarity of the final state to the target reference.

\item[\faGlasses] \textbf{Scene Consistency} \((\mathcal{R}_{\text{env}})\) measures consistency with the initial scene, penalizing drift, misplaced objects, or transitions into unrelated environments.

\item[\faGlasses] \textbf{Temporal Causality} \((\mathcal{R}_{\text{temp}})\) assesses whether the motion evolves in a physically plausible, coherent, and causal manner without temporal artifacts.

\item[\faGlasses] \textbf{Perceptual Fidelity} \((\mathcal{R}_{\text{per}})\) captures visual clarity and absence of distortions or artifacts.
\end{itemize}
Rewards are combined into a single total reward for each rollout, $\mathcal{R}_{\text{total}}(\vx^{(k)}_{1:T},c)$. We use strong vision–language models as non-parametric evaluators to score the generated videos. Detailed reward definitions and complete evaluation prompts are provided in Appendices~\ref{sec:VDNFTDetails} and~\ref{app:prompts}, respectively.

\section{Experiments}

\begin{table*}[t!]
    \centering
    \setlength{\tabcolsep}{8pt}
    \caption{\textbf{Quantitative comparisons on the \datasetname{} benchmark.} \modelnamenc{} outperforms all baselines across semantic, perceptual, and temporal metrics.}
    \label{tab:main_table}
    \resizebox{\textwidth}{!}{%
    \begin{tabular}{lcccccccc}
        \toprule
\rowcolor{TableHeader}
        \textbf{Model} & \textbf{DINO-Score}\(\uparrow\) & \textbf{CLIP-Score}\(\uparrow\) & \textbf{SSIM} \(\uparrow\)  & \textbf{LPIPS} \(\downarrow\) & \textbf{FVD} \(\downarrow\) & \textbf{flow MSE} \(\downarrow\) & \textbf{PSNR} \(\uparrow\) \\
        \midrule
        \textbf{EgoDreamer}~\citep{wang2024egovid} & 42.35 & 25.40 &    0.58 & 0.35 & 580.45 & 8.15 & 15.20  \\    
        \textbf{Handi}~\citep{li2024handi} & 31.12 & 18.25 & 0.42 & 0.52 & 912.30 & 14.50 & 12.85  \\
        \textbf{Cosmos}~\citep{nvidia2025worldsimulationvideofoundation} & 49.42 & 29.77 &    0.70   & 0.26 & 448.12 & 6.40 & 18.73  \\    
        \textbf{HunyuanVideo}~\citep{kong2024hunyuanvideo} &   53.54 & 29.43 &  0.71    & 0.26 & 384.31 & 6.10 & 18.88  \\      
        \textbf{WAN2.2}~\citep{wan2025wan}    & 53.99 & 35.69 & 0.72    & 0.23 & 322.17 & 5.78 & 20.44 \\        
         \rowcolor{TableHighlight} \textbf{\modelname{} (Ours)}    & \textbf{61.25} & \textbf{39.30} & \textbf{0.79}  & \textbf{0.15} & \textbf{182.25} & \textbf{2.83} & \textbf{24.08} \\     
        \bottomrule
    \end{tabular}
    }
\end{table*}

\begin{table*}[t!]
    \centering
        \setlength{\tabcolsep}{8pt}
    \caption{\textbf{Quantitative comparisons on \datasetname{} between \modelnamenc{} and other finetuned baseline variants.} +EV: with exo-view img, +TT: text-only domain adaptation; +CI: our conditioning inputs  (exo-view and goal instructions) with our structured injection with Geometry Weak Supervision.}
    \label{tab:main_table2}
    \resizebox{0.99\linewidth}{!}{%
    \begin{tabular}{lcccccccc}
        \toprule
\rowcolor{TableHeader}
        \textbf{Model} & \textbf{DINO-Score}\(\uparrow\) & \textbf{CLIP-Score}\(\uparrow\) & \textbf{SSIM} \(\uparrow\)  & \textbf{LPIPS} \(\downarrow\) & \textbf{FVD} \(\downarrow\) & \textbf{flow MSE} \(\downarrow\) & \textbf{PSNR} \(\uparrow\) \\
        \midrule
        \textbf{Cosmos+EV} & 48.60 & 29.60 & 0.67 & 0.28 & 485.75 & 6.82 & 18.30\\
        \textbf{Cosmos+TT} & 50.80 & 30.40 & 0.71 & 0.25 & 433.90 & 6.31 & 18.88\\
        \textbf{HunyuanVideo+EV} & 52.80 & 29.20 & 0.70 & 0.27 & 405.87 & 6.30 & 18.61\\
        \textbf{HunyuanVideo+TT} & 54.10 & 29.86 & 0.72 & 0.24 & 365.80 & 5.95 & 19.10\\
        \textbf{WAN2.2+EV} & 52.91 & 35.11 & 0.71 & 0.27 & 352.41 & 6.25 & 20.05\\
        \textbf{WAN2.2+TT} & 54.80 & 36.20 & 0.73 & 0.25 & 310.57 & 5.60 & 20.64\\
        \textbf{WAN2.2+CI} & 58.92 & 38.05 & 0.76 & 0.18 & 218.72 & 3.92 & 22.87\\
         \rowcolor{TableHighlight} \textbf{\modelname{} (Ours)}    & \textbf{61.25} & \textbf{39.30} & \textbf{0.79}  & \textbf{0.15} & \textbf{182.25} & \textbf{2.83} & \textbf{24.08} \\     
        \bottomrule
    \end{tabular}
    }
\end{table*}

\noindent
\textbf{\datasetname{} Benchmark.}
We introduce \datasetnamenc{}, the first large-scale benchmark designed to evaluate the ability of egocentric world models to synthesize complex scenes involving multiple, fine-grained conditioning signals. \datasetnamenc{} is curated from the Nymeria ~\citep{ma2024nymeria} and Ego-Exo4D ~\citep{grauman2024ego} datasets. To better understand human–environment interaction, we further enrich this benchmark with dense annotations of hand–object dynamics, object state changes, and step-level action semantics. The corpus contains 15,000 training samples and a separate, non-overlapping set of 100 held-out test clips covering all interaction categories in our taxonomy. Details about the annotation and dataset statistics are provided in Appendix~\ref{sec:bench_detail}. Caption and action-segmentation prompts appear in Appendix~\ref{app:prompts}.

\noindent \textbf{Implementation Details and Metrics}
We provide implementation details in Appendix \ref{sec:implement}. For evaluation, we measure low-level visual fidelity using PSNR~\citep{huynh2008scope} and SSIM~\citep{wang2004image}. To assess perceptual alignment, we employ LPIPS~\citep{zhang2018perceptual}, DINO-Score~\citep{zhang2022dino} and CLIP-Score~\citep{radford2021learning}, which captures higher-level semantic correspondence. We further measure distributional realism using FVD~\citep{rakhimov2020latent}. More details are in Appendix \ref{sec:more_metrix}.

\subsection{Experimental Results}
\noindent \textbf{Quantitative comparisons.}
We compare \modelnamenc{} with state-of-the-art general-purpose video models, including Cosmos~\citep{nvidia2025worldsimulationvideofoundation}, HunyuanVideo~\citep{kong2024hunyuanvideo}, and WAN2.2~\citep{wan2025wan}, as well as models designed for egocentric generation such as EgoDreamer~\citep{wang2024egovid} and Handi~\citep{li2024handi}.  These models represent the strongest available general-purpose video generators and the most relevant publicly available egocentric baselines. To ensure a fair evaluation, we fine-tune video models on our \datasetnamenc{} dataset to bridge the domain gap.\looseness-1

As shown in Table~\ref{tab:main_table}, \modelname{} consistently outperforms all baselines across all metrics. Compared with the strongest baseline, it improves semantic alignment by +13.4\% DINO-Score and +10.1\% CLIP-Score, while increasing structural fidelity (+9.7\% SSIM) and reconstruction quality (+17.8\% PSNR). At the same time, perceptual error is substantially reduced (35\% lower LPIPS). Most notably, \modelnamenc{} achieves large gains in temporal modeling, reducing FVD by 43\% and flow MSE by 51\%, indicating significantly more coherent motion and stable scene dynamics. 
These improvements demonstrate that \modelnamenc{} more effectively captures egocentric motion patterns than baselines.

To ensure a fair and rigorous comparison, we further enhance the baseline models using three progressive strategies and compare under identical training settings. Native exo-image conditioning (+EV) supplies the auxiliary exocentric image through each backbone's existing visual-conditioning interface, then fine-tunes the model on \datasetnamenc{} with its standard denoising objective. Goal-text domain fine-tuning (+TT) fine-tunes the backbone on the same \datasetnamenc{} videos through its native text-conditioning pathway, using the high-level goal instruction and no exocentric image. Structured conditioning injection with geometry weak supervision (+CI) replaces both pathways with our conditioning mechanism and geometry supervision.
As shown in Table~\ref{tab:main_table2}, both baseline strategies help, and +EV performs worse than +TT on all three backbones, which shows that supplying the exocentric image through a native interface is not by itself sufficient. \modelnamenc{} achieves the best results across all seven metrics, reaching an FVD of $182.25$ and a flow MSE of $2.83$.

\noindent \textbf{Human evaluation.} We conduct a human study to assess quality and goal alignment; the protocol and results, including the matched Stage-1 comparison are reported in Appendix~\ref{app:user}, Tables~\ref{tab:user_study} and~\ref{tab:user_study_stage1}.

\subsection{Ablations}

\begin{table*}[t!]
    \centering
        \setlength{\tabcolsep}{5pt}
    \caption{\textbf{Ablation on \modelnamenc{} modules.} We evaluate the impact of diffusion finetuning (FT), geometry weak supervision (GWS), and the proposed \graft{} refinement. Each component consistently improves performance across semantic, perceptual, and temporal metrics, with the full \modelnamenc{} model achieving the best overall results.}
    \label{tab:ablation_table}
    \resizebox{0.99\textwidth}{!}{%
    \begin{tabular}{cccccccccc}
        \toprule
\rowcolor{TableHeader}
        \textbf{FT} & \textbf{GWS} & \textbf{\graft{}} 
        & \textbf{DINO-Score}\(\uparrow\) 
        & \textbf{CLIP-Score}\(\uparrow\) 
        & \textbf{SSIM}\(\uparrow\)  
        & \textbf{LPIPS}\(\downarrow\) 
        & \textbf{FVD}\(\downarrow\) 
        & \textbf{flow MSE}\(\downarrow\) 
        & \textbf{PSNR}\(\uparrow\) \\
        \midrule    

        \cmark & \xmark & \xmark 
        & 56.81 & 37.10 & 0.74 & 0.21 & 260.89 & 4.82 & 21.92 \\      

        \cmark & \cmark & \xmark 
        & 58.92 & 38.05 & 0.76 & 0.18 & 218.72 & 3.92 & 22.87 \\    

        \rowcolor{TableHighlight} \cmark & \cmark & \cmark & \textbf{61.25} & \textbf{39.30} & \textbf{0.79} 
& \textbf{0.15} & \textbf{182.25} & \textbf{2.83} & \textbf{24.08} \\
        \bottomrule
    \end{tabular}
    }
\end{table*}

\noindent \textbf{Effectiveness of \graft{}.} Table~\ref{tab:ablation_table} isolates reward-guided refinement by comparing FT+GWS with the full FT+GWS+\graft{} model, keeping conditioning and geometry supervision fixed. Adding \graft{} increases DINO-Score from 58.92 to 61.25 and CLIP-Score from 38.05 to 39.30, while reducing FVD from 218.72 to 182.25 and flow MSE from 3.92 to 2.83. These gains support the effectiveness of trajectory-level refinement for semantic alignment and temporal coherence. A naive per-frame adaptation of an image-level reward objective is worse than no refinement at all (FVD 231.60 vs. 218.72, Appendix~\ref{app:graft_design}, Table~\ref{tab:ablation_nft}).

\noindent \textbf{Effect of Geometric Alignment Loss.} We compare FT with FT+GWS in Table~\ref{tab:ablation_table}, with \graft{} disabled in both variants. Geometry alignment increases DINO-Score by 2.11 (56.81 to 58.92) and CLIP-Score by 0.95 (37.10 to 38.05), while reducing LPIPS from 0.21 to 0.18 and flow MSE from 4.82 to 3.92. Appendix~\ref{app:components}, Table~\ref{tab:ablation_components}, shows complementary gains from geometry supervision and \graft{}, while extra denoising fine-tuning remains below the full model. Appendix~\ref{app:geometry}, Table~\ref{tab:ablation_gws}, shows that geometric alignment outperforms semantic and random feature alignment.
\begin{table*}[t]
  \centering
  \renewcommand{\arraystretch}{1.10}
  \setlength{\tabcolsep}{4.4pt}
  \caption{\textbf{Ablation of conditioning inputs.}
  Missing modalities are replaced with learned null embeddings. 
  For counterfactual conditions, \emph{random Exo} uses an exocentric view sampled from another test example, and \emph{wrong Text} uses an action-incompatible instruction.
  Metrics computed against original held-out video and instruction.\looseness-1}
  \label{tab:conditioning_ablation}

  \resizebox{\textwidth}{!}{%
  \begin{tabular}{@{}lccc ccccccc@{}}
    \toprule
\rowcolor{TableHeader}
    \textbf{Conditioning}
    & \multicolumn{3}{c}{\textbf{Inputs}}
    & \multicolumn{7}{c}{\textbf{Generation Quality}} \\
    \cmidrule(lr){2-4}
    \cmidrule(lr){5-11}
    \rowcolor{TableHeader}
    & \textbf{Ego}
    & \textbf{Text}
    & \textbf{Exo}
    & \textbf{DINO} $\uparrow$
    & \textbf{CLIP} $\uparrow$
    & \textbf{SSIM} $\uparrow$
    & \textbf{LPIPS} $\downarrow$
    & \textbf{FVD} $\downarrow$
    & \textbf{Flow MSE} $\downarrow$
    & \textbf{PSNR} $\uparrow$ \\
    \midrule

    \rowcolor{TableGroup}
    \multicolumn{11}{@{}l}{\textbf{\textit{Single-modality conditioning}}} \\
    Text only
    & \xmark & \cmark & \xmark
    & 49.84 & 36.75 & 0.62 & 0.33 & 398.52 & 7.01 & 17.18 \\

    Ego only
    & \cmark & \xmark & \xmark
    & 54.26 & 28.41 & 0.70 & 0.25 & 310.44 & 5.73 & 20.14 \\

    Exo only
    & \xmark & \xmark & \cmark
    & 50.91 & 27.86 & 0.65 & 0.30 & 365.73 & 6.38 & 18.56 \\

    \midrule
    \rowcolor{TableGroup}
    \multicolumn{11}{@{}l}{\textbf{\textit{Two-modality conditioning}}} \\
    Ego + Text
    & \cmark & \cmark & \xmark
    & 58.73 & 38.72 & 0.77 & 0.17 & 204.62 & 3.19 & 23.30 \\

    Ego + Exo
    & \cmark & \xmark & \cmark
    & 56.88 & 30.95 & 0.76 & 0.18 & 231.48 & 3.95 & 22.56 \\

    Text + Exo
    & \xmark & \cmark & \cmark
    & 55.64 & 37.91 & 0.69 & 0.25 & 282.36 & 4.91 & 19.69 \\

    \midrule
    \rowcolor{EgoPink!35}
    \multicolumn{11}{@{}l}{\textbf{\textit{Counterfactual conditioning}}} \\
    Ego + Text + random Exo
    & \cmark & \cmark & rand.
    & 57.20 & 37.89 & 0.74 & 0.21 & 237.44 & 4.13 & 21.85 \\

    Ego + wrong Text + Exo
    & \cmark & wrong & \cmark
    & 55.31 & 31.68 & 0.75 & 0.20 & 254.18 & 4.55 & 21.84 \\

    \midrule
    \rowcolor{TableHighlight}
    \textbf{Ego + Text + Exo}
    & \cmark & \cmark & \cmark
    & \best{61.25}
    & \best{39.30}
    & \best{0.79}
    & \best{0.15}
    & \best{182.25}
    & \best{2.83}
    & \best{24.08} \\
    \bottomrule
  \end{tabular}%
  }

\end{table*}

\begin{figure*}[!t]
\vspace{0.3cm}
\centering
    \includegraphics[width=0.99\textwidth]{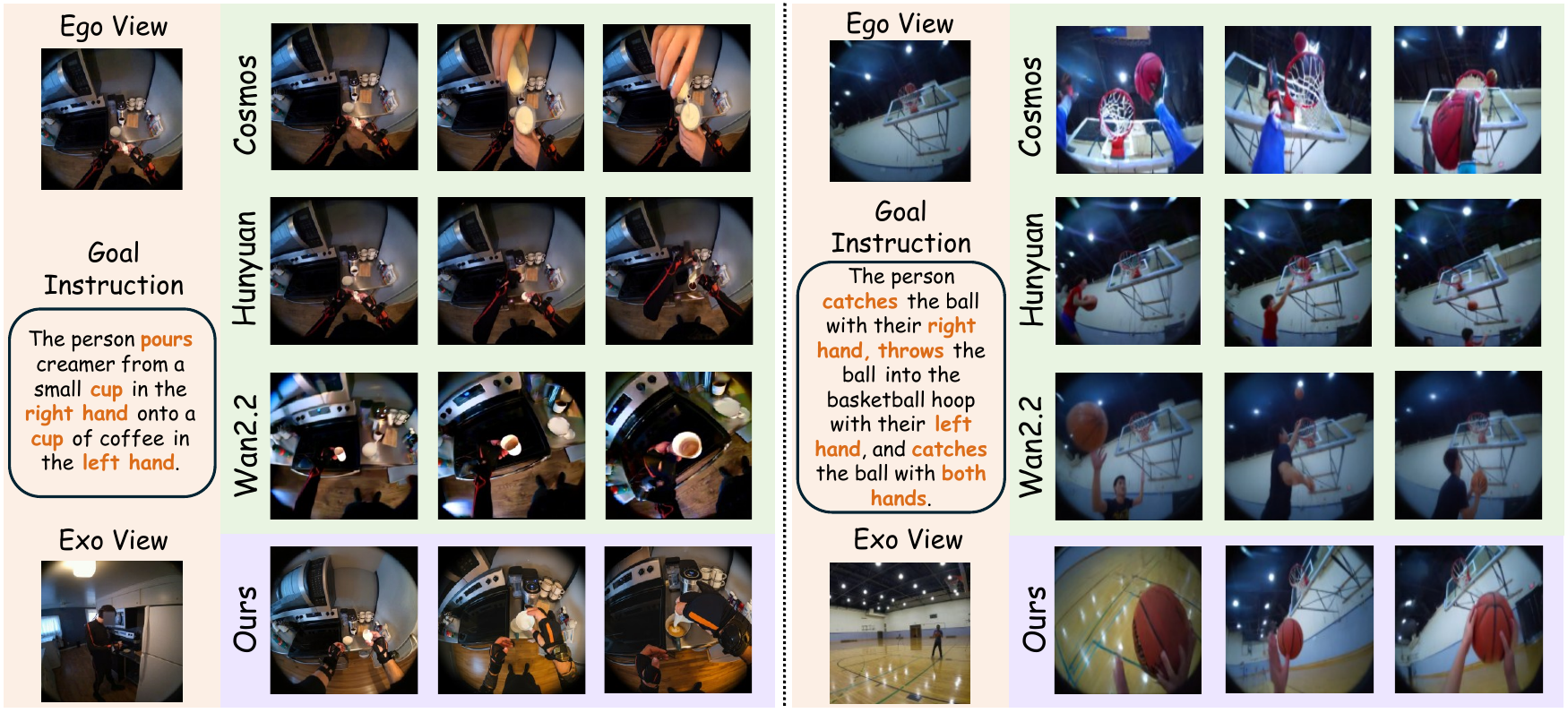}
\captionof{figure}{\textbf{Qualitative Comparison between \modelname{} and baselines}. \modelnamenc{}  accurately reconstructs multi-step, causally ordered actions, preserving hand–object geometry, temporal consistency, and goal alignment. For instance, in the first example, Cosmos erroneously generates a third hand, Hunyuan depicts a disconnected arm, and Wan2.2 fails to complete the coffee-pouring task. In the second example, Cosmos generates multiple balls, while both Hunyuan and Wan2.2 generate an incorrect person to perform the action. In contrast, \modelnamenc{} accurately completes both tasks.}
\label{fig:quali}
\end{figure*}

\noindent \textbf{Conditioning inputs.}
Table~\ref{tab:conditioning_ablation} studies whether \modelnamenc{} uses each conditioning signal: the egocentric observation, the goal instruction, and the auxiliary exocentric image. Each missing modality is replaced with a learned null embedding. Ego-only generation preserves more low-level scene structure but lacks goal alignment. Text-only generation achieves better text alignment but suffers from poor visual and temporal fidelity because it lacks the first-person initialization. Ego+Text already produces plausible rollouts, showing that the exocentric view is optional at inference time. Adding the correct exocentric reference further improves all metrics, reducing FVD from 204.62 to 182.25 and flow MSE from 3.19 to 2.83. We also evaluate two counterfactual settings. Replacing the paired exocentric image with a random exo-view hurts all metrics, indicating that \modelnamenc{} does not ignore the exocentric input. Replacing the instruction with an incorrect goal causes a large drop in CLIP-Score and temporal fidelity, showing that the rollout is sensitive to high-level intent rather than being determined only by visual context. 
The exocentric image used is taken from the end of the clip. Appendix~\ref{app:exo_timestamp}, Table~\ref{tab:exo_timestamp}, shows that a clip-start reference provides most of the benefit, while Appendix~\ref{app:clipstart}, Table~\ref{tab:clipstart_ablation}, repeats the component ablation under that setting. Appendix~\ref{app:instruction}, Table~\ref{tab:instruction_sensitivity}, shows that refinement increases sensitivity to the goal instruction. Finally, Appendix~\ref{app:exo_removed}, Table~\ref{tab:exo_deleted}, shows that refinement gains persist without exocentric input.

\noindent \textbf{Reward Ablation.} Ablating $\mathcal{R}_{\text{goal}}$, $\mathcal{R}_{\text{env}}$, $\mathcal{R}_{\text{temp}}$, or $\mathcal{R}_{\text{per}}$ degrades performance, with the full reward composition performing best across all reported metrics (Appendix~\ref{app:reward_ablation}, Table~\ref{tab:reward_ablation}). Appendix~\ref{app:reward_reliability} evaluates cross-judge agreement and reliability (Tables~\ref{tab:reward_cross_vlm} and~\ref{tab:reward_icc}), robustness to judge replacement, agreement with human ratings (Table~\ref{tab:reward_human}), and reward-hacking probes.

\label{sec:quali}

\begin{figure}[t!]
  \centering
  \includegraphics[width=0.99\textwidth]{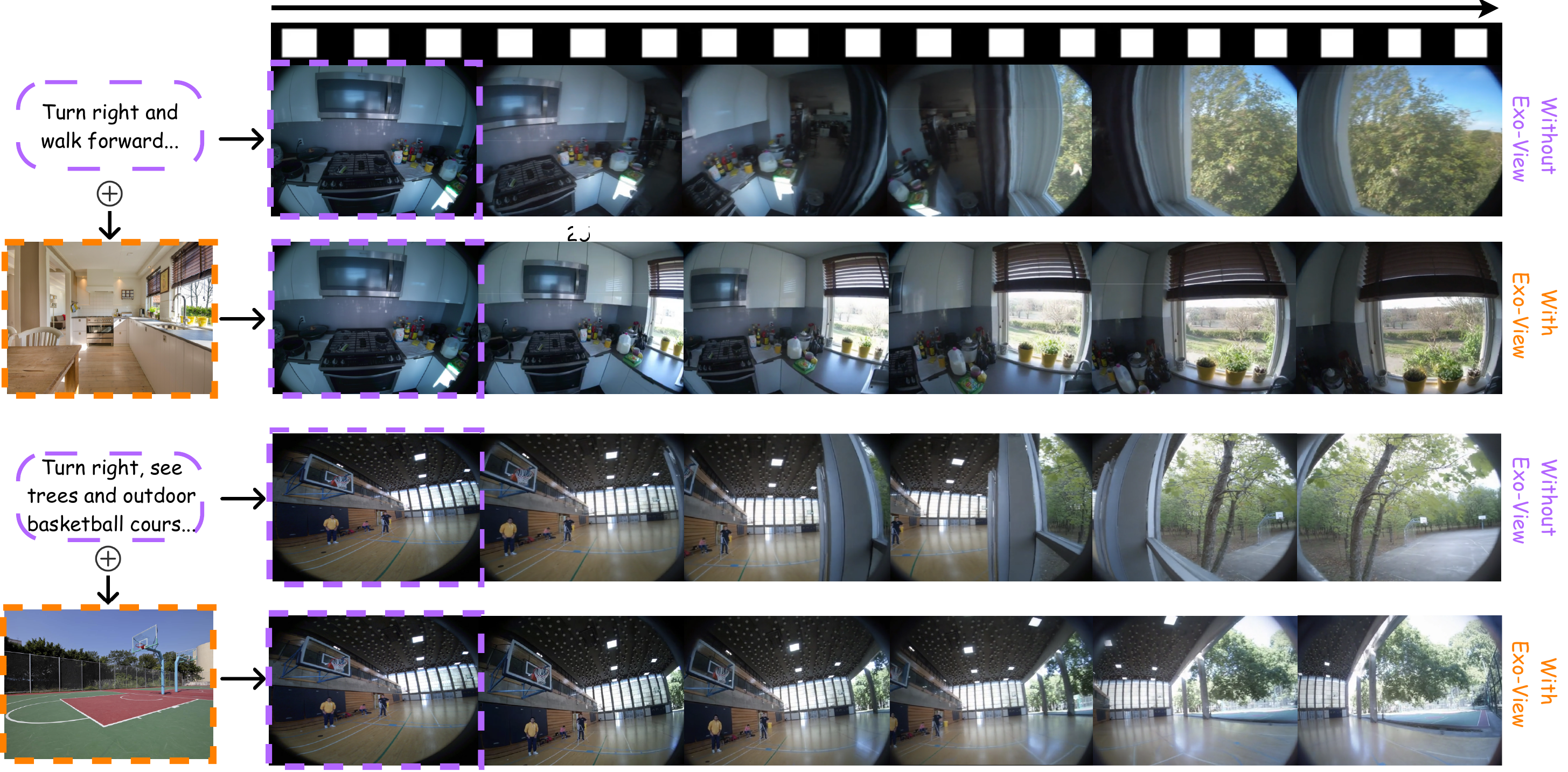}
 \caption{\textbf{Qualitative comparison with and without exocentric input.}   \modelnamenc{} can be steered using an auxiliary exo-view to improve spatial grounding by anchoring the simulation to the reference environment.
  }
  \label{fig:simulation}
\end{figure}

\subsection{Qualitative Analysis} 
Figure \ref{fig:quali} shows comparisons of our model, \modelnamenc{}, against Cosmos~\citep{agarwal2025cosmos}, Hunyuan~\citep{kong2024hunyuanvideo}, and Wan2.2~\citep{wan2025wan} for two complex, long-horizon egocentric tasks. Results demonstrate that \modelnamenc{} generates more plausible and coherent video sequences than the baseline models. Additional examples are available in Appendix \ref{sec:vis_results}.

Figure~\ref{fig:simulation} provides a qualitative comparison of \modelnamenc{}'s generation with and without an auxiliary exocentric image. Results show that \modelnamenc{} generates plausible egocentric rollouts using instruction guidance alone  (Rows 1 \& 3), while incorporating an auxiliary exocentric image further improves spatial grounding and scene consistency by anchoring the simulation to the reference environment (Rows 2 \& 4), such as the \textit{potted plants on the windowsill} and the distinctive \textit{red and green rubberized surface} of the court. This confirms that an exo-view frame can serve as effective guidance for \modelnamenc{}. Dense rollout visualizations across diverse activities are provided in Appendix~\ref{sec:vis_results}, Figure~\ref{fig:rebuttal_fig}, alongside additional baseline comparisons and long-duration sequences (Figures~\ref{fig:quali2} and~\ref{fig:long_sequence_appendix}).\looseness-1

\noindent \textbf{Real-world Smart-Glasses Experiments} Previous egocentric world models have been tested only with in-domain data, ignoring the complexity of real-world out-of-domain (OOD) data. To fill this gap, we deploy \modelnamenc{} in a real-world environment using ARGO smartglasses. The selected tasks are shown in Figure~\ref{fig:teaser}. For each task, a single egocentric frame is captured directly from the smartglasses, while an auxiliary exo-view reference image from another participant is provided alongside the text-action guidance. Details are provided in Appendix~\ref{sec:devices}, Figure~\ref{fig:glass}.
\section{Conclusion}
We introduce \modelnamenc{}, a goal-directed egocentric world simulator that generates coherent first-person video rollouts from minimal context: an egocentric observation, a high-level instruction, and an optional exocentric reference. \modelnamenc{} combines geometry-aware grounding for spatial consistency with \graft{}, a trajectory-level reward-guided refinement method that balances goal completion, scene stability, temporal causality, and perceptual fidelity through a negative-aware update of the diffusion velocity field. Experiments, including smart-glasses scenarios, show that \modelnamenc{} outperforms strong baselines, producing rollouts with stronger instruction alignment, more realistic egocentric motion, and more stable scene evolution. We hope this direction will facilitate future research on immersive world models, interactive simulation, and human-centered XR systems.

% ===================== Bibliography =====================
\bibliographystyle{plainnat}
\bibliography{egoforge}

% ===================== Appendix =====================
\appendix

\renewcommand{\topfraction}{0.95}
\renewcommand{\bottomfraction}{0.9}
\renewcommand{\textfraction}{0.05}
\renewcommand{\floatpagefraction}{0.85}
\clearpage

\section{Why Trajectory-Level Alignment}
\label{sec:appendix_complexity}
In the main paper, we introduce \graft{} as a trajectory-level reinforcement stage to guide generation using multiple, possibly conflicting conditioning signals. This appendix elaborates on why such an alignment is a non-trivial challenge, moving beyond standard per-frame conditioning. The difficulty arises from coordinating heterogeneous semantic and visual goals over entire, temporally extended video rollouts. We highlight three specific complexities that necessitate a trajectory-level optimization approach.\looseness-1

\noindent \textbf{Reward Sparsity and Temporal Credit Assignment.}
A primary challenge in egocentric simulation is that the criteria for success are often holistic and only measurable at an action sequence's conclusion. For instance, our goal completion reward ($\mathcal{R}_{\text{goal}}$) evaluates whether a multi-step task (\eg ``put the can back'') was \textit{ultimately} successful. This introduces a temporal credit assignment problem. If a generated trajectory of $T$ frames fails, it is non-trivial to pinpoint which frames or input-fusion errors contributed to the failure, and when those errors occurred during the rollout, contributed to the failure. Simple per-frame rewards (\eg pixel-wise losses) cannot capture this procedural, long-horizon objective. Therefore, a reinforcement learning mechanism is required to optimize for this sparse, trajectory-wide reward, propagating the global success signal back to the entire generative process.\looseness-1

\noindent \textbf{Preventing Goals Drift.}
In long-duration video synthesis, generative models are prone to ``goals drift,'' where the generated sequence gradually deviates from the initial conditioning signals. For example, while DiT blocks are conditioned on inputs at each step, there is no explicit mechanism in the standard diffusion objective to guarantee that a scene's geometric structure or background elements (part of $\mathcal{R}_{\text{env}}$) remain consistent from frame 1 to frame $T$. Without a trajectory-level regularizer, the model might ``forget'' or ``mutate'' these goals, leading to artifacts such as a drifting environment or object flickering. The \graft{} stage acts as this global supervisor, evaluating the entire rollout and penalizing sequences that exhibit such drift, thereby enforcing long-term temporal coherence.\looseness-1

\noindent \textbf{The "Shortcut" Problem in Multimodal Fusion.}
As we note in the main paper, heterogeneous conditioning signals may dominate, conflict, or cancel out. A significant challenge is that the model can learn "shortcuts" to minimize the training loss without effectively combining information from the conditioning inputs. For example, the model might find it easier to overfit to the auxiliary visual inputs, producing a stylistically similar video while entirely ignoring the complex procedural instructions in the text input. This results in a visually plausible rollout that fails the semantic task. Our multi-dimensional reward function (assessing goal, environment, causality, and fidelity) and the negative-aware finetuning structure address this problem. By evaluating rollouts across multiple axes and explicitly repelling the model from ``suboptimal'' samples (\eg low $\mathcal{R}_{\text{goal}}$ but high $\mathcal{R}_{\text{per}}$), \graft{} compels the model to find a policy that balances all inputs, discouraging shortcut solutions and enforcing the joint use of conditioning inputs.\looseness-1

\section{\graft{} Reward Details}
\label{sec:VDNFTDetails}
This appendix details reward design choices for our proposed \graft{}.\looseness-1

\noindent \textbf{Goal Completion Reward ($\mathcal{R}_{\text {goal}}$).}
The goal completion reward ($\mathcal{R}_{\text {goal}}$) evaluates the success of the manipulation task based on both the process and the final outcome. This 2.0-point metric is a composite function, $\mathcal{R}_{\text{goal}}\!=\!\mathcal{R}_{\text{task}}\!+\!\mathcal{R}_{\text{align}}$, where each component has a maximum value of 1.0. The Task Completion score ($\mathcal{R}_{\text{task}}$) assesses the dynamic fidelity of the video, rewarding the agentic world model for performing the state changes and manipulations required to transition from the initial state to the final one. Concurrently, the Visual Alignment score ($\mathcal{R}_{\text{align}}$) evaluates the target outcome of the action by measuring the semantic correspondence and spatial relationship of key elements between the video's final frame and the target state image, ensuring the end state is verifiably correct.\looseness-1

\noindent \textbf{Scene Consistency Reward ($\mathcal{R}_{\text{env}}$).}
To ensure the model focuses specifically on the manipulation task without altering the surrounding scene, we introduce the scene consistency reward ($\mathcal{R}_{\text {env}}$). This 2.0-point metric is critical for disentangling the required action from the background context, particularly when the target image depicts in a different environment. It is defined as $\mathcal{R}_{\text {env}}\!=\!\mathcal{R}_{\text{consist}}\!+\!\mathcal{R}_{\text{contam}}$. The Consistency Score ($\mathcal{R}_{\text{consist}}$) measures the temporal stability of static environmental features between the initial state image and the entire video sequence, penalizing any unmotivated environmental drift. The Contamination Score ($\mathcal{R}_{\text{contam}}$) specifically penalizes the model for "hallucinating" or leaking environmental features (\eg lighting, background objects, textures) that are present only in the target image but not in the initial state.\looseness-1

\noindent \textbf{Temporal Causality Reward ($\mathcal{R}_{\text {temp}}$).}
The temporal causality reward ($\mathcal{R}_{\text {temp}}$) assesses the physical and logical coherence of the generated video's dynamics. A video that successfully achieves the goal but does so through impossible or nonsensical motion is penalized. This 2.0-point metric, $\mathcal{R}_{\text {temp}}\!=\!\mathcal{R}_{\text{phys}}\!+\!\mathcal{R}_{\text{logic}}$, evaluates how plausibly the task is performed. The Physics Plausibility score ($\mathcal{R}_{\text{phys}}$) scrutinizes the sequence for adherence to real-world physical principles, such as momentum, gravity, and object permanence, penalizing unrealistic motion or object interactions. To complement this, the Causal Logic score ($\mathcal{R}_{\text{logic}}$) ensures that all state changes (effects) are preceded by logical and visible interactions (causes), such as an agent's hand grasping an object before it moves, thereby preventing spurious correlations or "action-at-a-distance" phenomena.

\noindent \textbf{Perceptual Fidelity Reward ($\mathcal{R}_{\text{per}}$).}
This score is computed as the sum of three clipped and normalized components, $\mathcal{R}_{\text{PSNR}}$, $\mathcal{R}_{\text{FVD}}$, and $\mathcal{R}_{\text{LPIPS}}$, each contributing a maximum of $\frac{2}{3}$ points. PSNR measures low-level, pixel-based reconstruction fidelity, LPIPS assesses perceptual similarity by comparing deep features in a manner aligned with human judgment, and FVD is a distributional metric that evaluates the overall realism of both visual appearance and temporal motion against a set of real videos.\looseness-1

\section{\datasetname{} Details}
\label{sec:bench_detail}
We introduce \datasetnamenc{}, the first large-scale dataset and benchmark designed to evaluate the capability of egocentric world models to synthesize complex, goal-directed scenes from sparse static context. \datasetnamenc{} is constructed from the Nymeria~\citep{ma2024nymeria} and Ego-Exo4D~\citep{grauman2024ego} datasets. We segment the videos based on the action annotations provided with these two datasets, resulting in clips that are each 10 seconds long. We then instruct an expert temporal-action summarizer to generate concise descriptions of stationary atomic actions lasting 10 seconds. The summarizer is specifically constrained to select hand-on-object manipulations while strictly avoiding non-stationary actions, locomotion, speech, and idle states, relying primarily on the provided filtered annotations to ensure the selected action dominates the segment. This setup guarantees that the synthesized descriptions focus exclusively on fine-grained in-place manipulation tasks.

To better understand human–environment interaction, we enrich this benchmark with dense annotations that detail fine-grained hand-object dynamics, object state changes, and step-level semantics. Instead of simple textual expansion, we instruct a multimodal language model to refine the concise atomic action description by grounding it in the actual video content, ensuring that the generated caption strictly adheres to the visual evidence. The output follows a structured four-sentence format, sequentially detailing the initial visual setup, the micro-dynamics of the motion, the physical reaction of the object, and the outcome. The dataset contains 15,000 training samples plus 100 separately held-out test clips, for a total of 15,100 samples. The test set is disjoint from the training set: none of its 100 clips is included in either the 13,000-sample Stage-1 training set or the 2,000-sample Stage-2 training set.
The test set contains samples that span the benchmark's interaction taxonomy, balancing broad coverage with the cost of evaluating long-horizon, high-fidelity video generation under multiple complementary metrics. 

\section{Implementation Details}
\label{sec:implement}
\noindent \textbf{Backbone and data.}
\modelnamenc{} uses the Wan2.2-5B~\citep{wan2025wan} model as the base generator. The 15,000 training samples are allocated to two stages; the separate 100-clip test set is used only for evaluation. We train with a batch size of 1 at a resolution of 720p, processing input videos at 24 fps and using 241 frames per sequence. Both stages fine-tune the diffusion transformer with Low-Rank Adaptation (LoRA) of rank 32, optimized with Adam at a learning rate of 1e\textsuperscript{-4} under mixed-precision (bf16) training on 8 H100 GPUs.

\noindent \textbf{Stage 1: denoising fine-tuning.}
The first stage fine-tunes the model on 13,000 samples with the velocity-matching loss $\mathcal{L}_{\text{D}}$ and the geometry weak supervision $\mathcal{L}_{\text{G}}$ of Section~\ref{sec:model}, sampling $t\sim\mathcal{U}(0,1)$ as in Eq.~\eqref{eq:forward}. The DINOv3 and VGGT backbones are frozen. This stage runs for 10 epochs and requires approximately 108 hours.

\noindent \textbf{Stage 2: \graft{}.}
The second stage uses the remaining 2,000 conditions. Only the diffusion transformer is trained, while all other components are frozen. The sampling policy $\pi^{\mathrm{old}}$ is the frozen Stage-1 model. For each condition we draw $K\!=\!6$ rollouts from $\pi^{\mathrm{old}}$ with the same sampler used at evaluation, giving 12,000 scored rollouts, and score each rollout with $\mathcal{R}_{\text{total}}$ (Appendix~\ref{sec:VDNFTDetails}). Only the clean rollouts and their rewards are stored; the sampling trajectories are not needed. Rewards are normalized within each group of $K$ rollouts: $\mu_c$ in Eq.~\eqref{eq:optimality} is the group mean and $Z_c$ is the standard deviation of the $K$ raw rewards, floored at a small constant to avoid division by zero. A condition whose rollouts all receive the same reward therefore maps every rollout to $r\!=\!\tfrac12$ and, by Eq.~\eqref{eq:nft_opt}, carries no reward signal: such groups only anchor $\boldsymbol{v}_\theta$ to the frozen $\boldsymbol{v}^{\text{old}}$. We set the guidance parameter to $\beta\!=\!0.1$ in Eq.~\eqref{eq:nft_loss}, sample $t\sim\mathcal{U}(0,1)$ and $\epsilon\sim\mathcal{N}(\mathbf{0},\mathbf{I})$ exactly as in Stage 1, and evaluate $\boldsymbol{v}^{\text{old}}$ and $\boldsymbol{v}_\theta$ on the same noisy latent $\vz_t$ so that the implicit positive and negative policies share one forward pass of the frozen model.\looseness-1

\section{Video Generation Evaluation}
\label{sec:more_metrix}
As summarized in the main paper, we employ a comprehensive suite of metrics to evaluate the performance of \modelnamenc{} on the X-Ego benchmark. We describe how each metric is computed below.. For semantic and perceptual alignment, we compute the \textbf{DINO-Score} to evaluate frame-level semantic similarity, which is calculated as the average cosine similarity between the DINOv2 ViT-g embeddings of the predicted video frames and the ground-truth video frames. We also compute the \textbf{CLIP-Score} to measure text-video alignment. This is calculated as the cosine similarity between the CLIP embedding of the generated video frames and the embedding of the corresponding textual guidance. Additionally, \textbf{LPIPS} (Learned Perceptual Image Patch Similarity) assesses perceptual distance by computing the $L_2$ distance between deep features extracted from a pretrained network (\eg VGG), aligning more closely with human judgments of visual similarity than pixel-based metrics. For low-level visual fidelity, \textbf{PSNR (Peak Signal-to-Noise Ratio)} measures pixel-level reconstruction quality by comparing the maximum possible pixel value to the Mean Squared Error (MSE) between the generated and ground-truth frames, where a higher PSNR indicates lower pixel-wise error. \textbf{SSIM (Structural Similarity Index)} assesses visual fidelity by comparing three components: luminance, contrast, and structural information between the generated and ground-truth frames. Finally, for distributional and temporal fidelity, \textbf{FVD (Fréchet Video Distance)} assesses distributional realism by measuring the Fréchet Inception Distance between distributions of real and generated videos. Features are extracted from a pretrained video classification model, and the distance between the two multivariate Gaussian distributions (one for real videos, one for generated) is computed. \textbf{Flow MSE} measures temporal and motion fidelity. We first compute the optical flow fields for both the generated and ground-truth video sequences, and then calculate the mean squared error (MSE) between the two sets of flow fields, penalizing discrepancies in predicted motion.\looseness-1

\section{Details of User Study}
\label{app:user}
We recruit 20 annotators to evaluate 25 groups of videos, where each group contains outputs from all competing methods paired with the corresponding text prompt. Participation was voluntary, and no monetary or non-monetary compensation was provided. Participants were provided with a detailed annotation protocol to rate each video on a 1-to-5 scale across five dimensions: Quality (overall visual coherence), Fidelity (identity preservation and absence of artifacts or distortions), Smooth Motion (temporal consistency and fluidity of subject movement), Smooth Environment (background stability and smoothness), and Alignment (semantic correspondence between video and input). As shown in Table~\ref{tab:user_study}, our method achieves substantial gains in Alignment (4.75) and Fidelity (4.71), surpassing the strongest competitors (\eg +1.60 in Alignment over WAN2.2). These results, combined with superior scores in temporal and environmental smoothness, validate the effectiveness of our approach in generating high-quality, prompt-aligned videos. 

\begin{table}[t!]
    \centering
    \setlength{\tabcolsep}{5pt}
    \small
    \caption{\textbf{User study.} Each rating dimension targets one property of the rollout. Q. is Quality (overall synthesis quality), F. is Fidelity (object consistency), M. is Smooth Motion (physical plausibility), E. is Smooth Environment (scene consistency), and A. is Alignment (goal completion). Each dimension is rated from 1 (worst) to 5 (best). *Best model variants from Table~\ref{tab:main_table2}.}
    \label{tab:user_study}
    \begin{adjustbox}{max width=\linewidth}

    \begin{tabular}{lccccc}
        \toprule
\rowcolor{TableHeader}
        \textbf{Model} & \textbf{Q.}$\uparrow$ & \textbf{F.}$\uparrow$ & \textbf{M.}$\uparrow$ & \textbf{E.}$\uparrow$ & \textbf{A.}$\uparrow$ \\
        \midrule
        \textbf{Cosmos*}~\citep{nvidia2025worldsimulationvideofoundation} & 3.29 & 2.54 & 3.07 & 2.47 & 2.19 \\
        \textbf{Hunyuan*}~\citep{kong2024hunyuanvideo} & 3.46 & 2.86 & 3.72 & 3.16 & 3.08 \\
        \textbf{WAN2.2*}~\citep{wan2025wan} & 3.22 & 3.48 & 3.82 & 4.07 & 3.15 \\
        \rowcolor{TableHighlight}\textbf{\modelname{} (Ours)} & \textbf{4.58} & \textbf{4.71} & \textbf{4.25} & \textbf{4.48} & \textbf{4.75} \\
        \bottomrule
    \end{tabular}%
    \end{adjustbox}
\end{table}

Table~\ref{tab:user_study_stage1} compares EgoForge before and after
\graft{} refinement. The Stage-1 model already includes our conditioning mechanism and geometry weak supervision, so this comparison isolates the contribution of \graft{}.
Human ratings improve across all four reported dimensions.
The largest gain is in alignment with the goal instruction,
which increases from 3.94 to 4.75, followed by fidelity
(4.16 to 4.71), environment stability (4.21 to 4.48), and
motion smoothness (4.02 to 4.25).
These results show that \graft{} improves instruction following
while also producing more consistent objects, more stable
backgrounds, and smoother motion.\looseness-1

\begin{table}[t!]
\centering
\small
\setlength{\tabcolsep}{6pt}
\renewcommand{\arraystretch}{1.15}
\caption{\textbf{Effect of \graft{} on human ratings.}
Stage-1 (+CI) uses the same conditioning mechanism and geometry
supervision as EgoForge but has not undergone \graft{} refinement.
Scores range from 1 (worst) to 5 (best), and $\Delta$ denotes
the improvement over Stage-1.
\graft{} improves all four reported dimensions, with the
largest gain in alignment with the goal instruction.}
\label{tab:user_study_stage1}
\begin{adjustbox}{max width=\linewidth}
\begin{tabular}{@{}llccc c@{}}
\toprule
\rowcolor{TableHeader}
\textbf{Rating dimension} & \textbf{Property measured}
& \textbf{Wan2.2*} & \textbf{Stage-1 (+CI)} & \textbf{\modelnamenc{}}
& \textbf{$\Delta$ over Stage-1} \\
\midrule
Alignment          & Goal completion       & 3.15 & 3.94 & \textbf{4.75} & $+0.81$ \\
Fidelity           & Object consistency    & 3.48 & 4.16 & \textbf{4.71} & $+0.55$ \\
Smooth Environment & Scene consistency     & 4.07 & 4.21 & \textbf{4.48} & $+0.27$ \\
Smooth Motion      & Physical plausibility & 3.82 & 4.02 & \textbf{4.25} & $+0.23$ \\
\bottomrule
\end{tabular}
\end{adjustbox}
\end{table}

Under identical conditioning and geometry supervision, \graft{} gives the largest gain in goal completion, and also improves object consistency, scene consistency, and physical plausibility. This ordering matches the design of the reward, where the goal term carries the largest weight in distinguishing successful from failed rollouts.\looseness-1

\begin{planbox}[EgoPeriwinkle]{User Study: Participant Instructions}
The following instructions were provided to participants:

“You will be shown a text prompt and several generated egocentric videos corresponding to the same prompt. Please evaluate each video independently. You may replay the videos as needed. For each video, assign a score from 1 (worst) to 5 (best) for each of the following criteria:

1. Quality: overall visual quality and coherence of the generated video. A score of 1 indicates severe visual artifacts or an unrealistic video; a score of 5 indicates a visually coherent and realistic video.

2. Fidelity: preservation of object identity and absence of artifacts or distortions. A score of 1 indicates severe object deformation, identity inconsistency, or hallucinated artifacts; a score of 5 indicates stable object appearance and few or no artifacts.

3. Smooth Motion: temporal consistency and smoothness of the subject or hand/object motion. A score of 1 indicates jittery, discontinuous, or physically implausible motion; a score of 5 indicates smooth and temporally coherent motion.

4. Smooth Environment: stability and smoothness of the background environment. A score of 1 indicates severe background flickering, drifting, or scene changes; a score of 5 indicates a stable and consistent environment.

5. Alignment: semantic correspondence between the generated video and the input text prompt. A score of 1 indicates that the video is unrelated to or fails to follow the prompt; a score of 5 indicates that the video clearly follows and completes the described action.

Please use the full 1-to-5 scale when appropriate and base your judgment only on the displayed prompt and video content.”

\end{planbox}

The study used a simple rating form displaying the text prompt, generated videos, and the five rating questions above. No additional custom interface was used.
The user study involved only anonymous subjective ratings of generated videos and text prompts. Participants were informed of the evaluation purpose, the rating criteria, and that their responses would be analyzed only in aggregate. No sensitive personal information was collected. The potential risks were minimal and limited to ordinary visual fatigue from viewing short videos. 

\section{Additional Ablation Studies}
\label{app:more_ablation}
\subsection{Component Factorization}
\label{app:components}
We first ablate the effects of denoising fine-tuning, geometry weak supervision (GWS), and \graft{}. The structured ego/exo/text conditioning pathway is kept fixed for all fine-tuned variants. We also include an extra-SFT control that uses the same additional optimization budget as \graft{} but continues training only with the standard denoising objective.
\begin{table*}[h]
\centering
\caption{
\textbf{Component ablation.}
GWS and \graft{} provide complementary gains. Extra SFT uses the same additional
training budget as \graft{} but does not use reward-guided refinement.
}
\label{tab:ablation_components}
\resizebox{\textwidth}{!}{
\begin{adjustbox}{max width=\linewidth}
\begin{tabular}{lccccccc}
\toprule
\rowcolor{TableHeader}
Variant
& DINO$\uparrow$ & CLIP$\uparrow$ & SSIM$\uparrow$ & LPIPS$\downarrow$
& FVD$\downarrow$ & Flow MSE$\downarrow$ & PSNR$\uparrow$ \\
\midrule
Denoising FT
& 56.81 & 37.10 & 0.74 & 0.21 & 260.89 & 4.82 & 21.92 \\
Denoising FT + GWS
& 58.92 & 38.05 & 0.76 & 0.18 & 218.72 & 3.92 & 22.87 \\
Denoising FT + \graft{}
& 59.46 & 38.61 & 0.77 & 0.17 & 203.84 & 3.41 & 23.35 \\
Denoising FT + GWS + extra SFT
& 59.18 & 38.22 & 0.77 & 0.17 & 211.36 & 3.74 & 23.14 \\
\rowcolor{TableHighlight}\textbf{\modelnamenc{}}
& \textbf{61.25} & \textbf{39.30} & \textbf{0.79} & \textbf{0.15}
& \textbf{182.25} & \textbf{2.83} & \textbf{24.08} \\
\bottomrule
\end{tabular}
\end{adjustbox}
}
\end{table*}
As shown in Table~\ref{tab:ablation_components}, denoising fine-tuning alone improves the base generator, but still leaves a large gap in temporal coherence and structural fidelity. Adding GWS reduces FVD from 260.89 to 218.72 and Flow MSE from 4.82 to 3.92, indicating better geometric and temporal stability. \graft{} also gives a strong improvement without GWS, reducing FVD to 203.84 and Flow MSE to 3.41. Combining both achieves the best results across all metrics. The extra-SFT control remains clearly below the full model, suggesting that the improvement is not simply due to additional training steps.

\subsection{\graft{} Design}
\label{app:graft_design}
Table~\ref{tab:gain_attribution} distinguishes the overall
improvement of \modelnamenc{} from the additional contribution
of \graft{}.
The full-system gain compares \modelnamenc{} with off-the-shelf
Wan2.2, capturing the combined effect of our conditioning
mechanism, denoising fine-tuning, geometry weak supervision,
and refinement.
The method-only gain compares \modelnamenc{} with Stage-1
(Wan2.2+CI), which already includes all components except
\graft{}.
Relative to off-the-shelf Wan2.2, the full system reduces FVD
by 43.4\% and Flow MSE by 51.0\%.
Relative to Stage-1, \graft{} provides further reductions
of 16.7\% and 27.8\%, respectively, while also improving
all five remaining metrics.\looseness-1

\begin{table}[t!]
\centering
\small
\setlength{\tabcolsep}{8pt}
\renewcommand{\arraystretch}{1.15}
\caption{\textbf{Gain attribution.} The left column compares \modelnamenc{} with the off-the-shelf Wan2.2 backbone. The right column compares the \graft{} stage with our equally conditioned Stage-1 model.\looseness-1}
\label{tab:gain_attribution}
\begin{adjustbox}{max width=\linewidth}
\begin{tabular}{@{}lll@{}}
\toprule
\rowcolor{TableHeader}
\textbf{Metric} & \textbf{Full-system gain} & \textbf{Method-only gain} \\
\midrule
DINO$\uparrow$     & $+13.4\%$ ($53.99 \to 61.25$)   & $+3.95\%$ ($58.92 \to 61.25$) \\
CLIP$\uparrow$     & $+10.1\%$ ($35.69 \to 39.30$)   & $+3.29\%$ ($38.05 \to 39.30$) \\
SSIM$\uparrow$     & $+9.7\%$ ($0.72 \to 0.79$)      & $+3.95\%$ ($0.76 \to 0.79$) \\
LPIPS$\downarrow$  & $-34.8\%$ ($0.23 \to 0.15$)     & $-16.7\%$ ($0.18 \to 0.15$) \\
FVD$\downarrow$    & $-43.4\%$ ($322.17 \to 182.25$) & $-16.7\%$ ($218.72 \to 182.25$) \\
Flow MSE$\downarrow$ & $-51.0\%$ ($5.78 \to 2.83$)   & $-27.8\%$ ($3.92 \to 2.83$) \\
PSNR$\uparrow$     & $+17.8\%$ ($20.44 \to 24.08$)   & $+5.3\%$ ($22.87 \to 24.08$) \\
\bottomrule
\end{tabular}
\end{adjustbox}
\end{table}

We next examine each design choice in \graft{}. All variants in
Table~\ref{tab:ablation_nft} start from the same Denoising FT + GWS checkpoint.
We compare against reward re-ranking, positive-only refinement, removing per-condition reward normalization, and a shuffled-reward control that breaks the correspondence between rollouts and their rewards.
Table~\ref{tab:ablation_nft} shows that full \graft{} outperforms
both per-frame refinement and inference-time reward re-ranking.
Applying the image-level DiffusionNFT objective independently to
video frames increases FVD from 218.72 to 231.60, performing worse
than the unrefined model. This result highlights the limitations
of frame-level evaluation: realistic individual frames do not
ensure that a video performs the required actions in the correct
order, maintains hand--object contact, or reaches the goal through
physically plausible transitions.
Best-of-6 reward re-ranking improves FVD to 196.44 but remains
worse than full \graft{} at 182.25, despite requiring six generated
candidates per condition at inference. Positive-only refinement
also improves over the unrefined model but falls short of the full
negative-aware objective, supporting the benefit of explicitly
discouraging low-reward trajectories rather than only reinforcing
high-reward ones. Performance also declines without per-condition
reward normalization, which accounts for differences in reward
scale across conditions. Shuffling the rewards across rollouts
raises FVD to 229.55, worse than no refinement, showing that the
gains depend on meaningful correspondence between rollouts and
their rewards.
Finally, removing future-reference reward terms retains most of
the improvement, achieving an FVD of 188.94. These terms therefore
provide an additional benefit, but \graft{} still substantially
improves generation without them.

\begin{table*}[t!]
\centering
\caption{\textbf{Ablation of \graft{} design choices.}
All variants start from the same Denoising FT + GWS checkpoint. Best-of-6 re-ranking selects the highest-reward rollout from six candidates at inference, without further training. Per-frame DiffusionNFT scores and refines individual frames
rather than complete rollouts.
Full \graft{} achieves the best performance across all reported
metrics, outperforming both alternatives and its ablated variants.
}
\label{tab:ablation_nft}
\resizebox{\textwidth}{!}{
\begin{adjustbox}{max width=\linewidth}
\begin{tabular}{lccccccc}
\toprule
\rowcolor{TableHeader}
Refinement method
& DINO$\uparrow$ & CLIP$\uparrow$ & SSIM$\uparrow$ & LPIPS$\downarrow$
& FVD$\downarrow$ & Flow MSE$\downarrow$ & PSNR$\uparrow$ \\
\midrule
No \graft{}
& 58.92 & 38.05 & 0.76 & 0.18 & 218.72 & 3.92 & 22.87 \\
Per-frame DiffusionNFT~\citep{zheng2025diffusionnftonlinediffusionreinforcement} (naive port)
& 58.13 & -- & -- & -- & 231.60 & 4.05 & -- \\
Shuffled-reward \graft{}
& 58.47 & 37.89 & 0.75 & 0.20 & 229.55 & 4.21 & 22.41 \\
Best-of-6 reward re-ranking
& 60.04 & 38.73 & 0.78 & 0.16 & 196.44 & 3.26 & 23.64 \\
Positive-only \graft{}
& 60.29 & 38.84 & 0.78 & 0.16 & 190.83 & 3.08 & 23.70 \\
\graft{} w/o per-condition normalization
& 60.12 & 38.70 & 0.77 & 0.17 & 194.38 & 3.18 & 23.56 \\
\graft{} w/o future-reference reward terms
& 60.72 & 39.04 & 0.78 & 0.16 & 188.94 & 3.02 & 23.77 \\
\rowcolor{TableHighlight}\textbf{Full \graft{}}
& \textbf{61.25} & \textbf{39.30} & \textbf{0.79} & \textbf{0.15}
& \textbf{182.25} & \textbf{2.83} & \textbf{24.08} \\
\bottomrule
\end{tabular}
\end{adjustbox}
}
\end{table*}

\subsection{Geometry Weak Supervision}
\label{app:geometry}
We further study the geometry alignment objective. All variants in
Table~\ref{tab:ablation_gws} use the same denoising fine-tuning and \graft{} stages, differing only in the auxiliary alignment target or loss form. Random feature alignment uses a frozen randomly initialized network with matched feature dimensionality. DINO feature alignment replaces VGGT features with semantic image features.

\begin{table*}[t!]
\centering
\caption{
\textbf{Ablation of geometry weak supervision.}
VGGT-based geometry alignment is more effective than random or semantic feature alignment.
Angular and scale consistency are complementary.
}
\label{tab:ablation_gws}
\resizebox{\textwidth}{!}{
\begin{adjustbox}{max width=\linewidth}
\begin{tabular}{lccccccc}
\toprule
\rowcolor{TableHeader}
Geometry variant
& DINO$\uparrow$ & CLIP$\uparrow$ & SSIM$\uparrow$ & LPIPS$\downarrow$
& FVD$\downarrow$ & Flow MSE$\downarrow$ & PSNR$\uparrow$ \\
\midrule
No GWS
& 59.46 & 38.61 & 0.77 & 0.17 & 203.84 & 3.41 & 23.35 \\
Random frozen feature alignment
& 58.74 & 38.02 & 0.75 & 0.20 & 207.92 & 3.58 & 22.60 \\
DINO feature alignment
& 59.88 & 38.76 & 0.77 & 0.17 & 198.70 & 3.30 & 23.43 \\
VGGT angular only
& 60.36 & 39.02 & 0.78 & 0.16 & 191.85 & 3.14 & 23.59 \\
VGGT scale only
& 60.08 & 38.86 & 0.77 & 0.16 & 196.44 & 3.27 & 23.47 \\
\rowcolor{TableHighlight}\textbf{VGGT angular + scale}
& \textbf{61.25} & \textbf{39.30} & \textbf{0.79} & \textbf{0.15}
& \textbf{182.25} & \textbf{2.83} & \textbf{24.08} \\
\bottomrule
\end{tabular}
\end{adjustbox}
}
\end{table*}

The random-feature control does not improve performance, showing that the gains are not caused by simply adding an auxiliary loss. DINO feature alignment brings moderate improvements, but remains weaker than VGGT-based alignment, especially on FVD and Flow MSE. This suggests that semantic feature alignment alone is insufficient for stable egocentric rollout generation. Between the two geometry losses, angular alignment provides stronger temporal gains, while scale alignment contributes to perceptual and structural stability. Their combination gives the best result across all metrics, supporting the role of
geometry-aware representation alignment in maintaining coherent first-person scene evolution.

\subsection{Instruction Sensitivity}
\label{app:instruction}
We evaluate how strongly the goal instruction influences the
generated rollout by replacing the correct instruction with one
specifying an incompatible action, while keeping the initial
egocentric frame fixed. In both settings, we compute CLIP-Score
against the original instruction. The resulting score decrease
measures how much changing the instruction reduces alignment with
the original goal, with a larger decrease indicating greater
instruction sensitivity.
As shown in Table~\ref{tab:instruction_sensitivity}, replacing the
instruction decreases CLIP-Score by only 1.09 points for
off-the-shelf Wan2.2 and 1.29 points for Wan2.2+TT, indicating
limited sensitivity to the instruction change. In contrast, the
decrease is 4.21 points for Stage-1 and 7.62 points for
\modelnamenc{}. The larger gap after \graft{} refinement indicates
that the generated rollouts are more responsive to changes in
the goal instruction.

\begin{table}[t!]
\centering
\small
\setlength{\tabcolsep}{8pt}
\renewcommand{\arraystretch}{1.15}
\caption{\textbf{Counterfactual instruction test.} CLIP-Score under the correct instruction and under an incompatible one, with the initial egocentric frame held fixed. A larger $\Delta$ means the model responds more strongly to the goal.}
\label{tab:instruction_sensitivity}
\begin{adjustbox}{max width=\linewidth}
\begin{tabular}{@{}lccc@{}}
\toprule
\rowcolor{TableHeader}
\textbf{Model} & \textbf{Correct instruction} & \textbf{Counterfactual instruction} & \textbf{$\Delta$}$\uparrow$ \\
\midrule
Wan2.2 off-the-shelf & 35.69 & 34.60 & 1.09 \\
Wan2.2+TT            & 36.20 & 34.91 & 1.29 \\
Wan2.2+CI (Stage-1)  & 38.05 & 33.84 & 4.21 \\
\rowcolor{TableHighlight}\textbf{\modelnamenc{}} & \textbf{39.30} & 31.68 & \textbf{7.62} \\
\bottomrule
\end{tabular}
\end{adjustbox}
\end{table}

\subsection{Evaluation without Exocentric Input}
\label{app:exo_removed}
We evaluate \modelnamenc{} without the auxiliary exocentric image
at inference, using only the initial egocentric frame and goal
instruction. Table~\ref{tab:exo_deleted} compares this setting
with the Stage-1 model, which retains all three conditioning inputs.
Despite receiving less visual context, \modelnamenc{} outperforms
Stage-1 on six of the seven metrics. FVD decreases from 218.72
to 204.62 and Flow MSE from 3.92 to 3.19, while CLIP-Score
increases from 38.05 to 38.72. DINO-Score is slightly lower
(58.73 vs.\ 58.92).
These results show that the refined model retains an advantage
over Stage-1 on most metrics even when the auxiliary view is
unavailable at inference.\looseness-1

\begin{table}[t!]
\centering
\small
\setlength{\tabcolsep}{6pt}
\renewcommand{\arraystretch}{1.15}
\caption{\textbf{Generation without an exocentric reference.}
Stage-1 (Wan2.2+CI) receives the initial egocentric frame, goal
instruction, and exocentric image, whereas \modelnamenc{} receives
only the egocentric frame and instruction at inference.
Despite the reduced visual input, \modelnamenc{} outperforms
Stage-1 on six of seven metrics.
$\Delta$ denotes the score for \modelnamenc{} minus the
corresponding Stage-1 score.\looseness-1}
\label{tab:exo_deleted}
\begin{adjustbox}{max width=\linewidth}
\begin{tabular}{@{}lccccccc@{}}
\toprule
\rowcolor{TableHeader}
\textbf{Config} & DINO$\uparrow$ & CLIP$\uparrow$ & SSIM$\uparrow$ & LPIPS$\downarrow$
& FVD$\downarrow$ & Flow MSE$\downarrow$ & PSNR$\uparrow$ \\
\midrule
Wan2.2+CI, with exo      & 58.92 & 38.05 & 0.76 & 0.18 & 218.72 & 3.92 & 22.87 \\
\rowcolor{TableHighlight}\modelnamenc{}, exo deleted & 58.73 & \textbf{38.72} & \textbf{0.77} & \textbf{0.17} & \textbf{204.62} & \textbf{3.19} & \textbf{23.30} \\
\midrule
$\Delta$ & $-0.19$ & $+0.67$ & $+0.01$ & $-0.01$ & $-14.10$ & $-0.73$ & $+0.43$ \\
\bottomrule
\end{tabular}
\end{adjustbox}
\end{table}

\subsection{Effect of Exocentric Reference Timestamp}
\label{app:exo_timestamp}
The exocentric reference used in the main results is taken from
the end of the clip and may therefore reveal the final object state.
To evaluate whether an auxiliary view remains useful without
providing a future frame, Table~\ref{tab:exo_timestamp} compares
clip-start and clip-end references with two controls: no exocentric
input and an exocentric image from an unrelated scene.
A clip-start reference improves all four reported metrics over
the no-reference setting, reducing FVD from 204.62 to 189.30
and Flow MSE from 3.19 to 2.94. Using a clip-end reference
further reduces these values to 182.25 and 2.83, respectively.
In contrast, supplying an image from an unrelated scene degrades
all four metrics relative to omitting the reference, indicating
that the generated rollout is sensitive to the reference content.
These results show that the auxiliary view provides a benefit
even when it depicts only the initial scene. The clip-end setting
achieves the strongest results, but differs from the no-reference
and clip-start settings by providing future visual information
at inference.

\begin{table}[t!]
\centering
\small
\setlength{\tabcolsep}{8pt}
\renewcommand{\arraystretch}{1.15}
\caption{\textbf{Effect of exocentric reference timing and content.}
We evaluate \modelnamenc{} with no exocentric input, a reference
from the start or end of the same clip, or an image from an
unrelated scene. A clip-start reference improves all four metrics
over no reference, while the clip-end reference used in the main
results provides further gains. An unrelated reference performs
worse than omitting the auxiliary view.}
\label{tab:exo_timestamp}
\begin{adjustbox}{max width=\linewidth}
\begin{tabular}{@{}lcccc@{}}
\toprule
\rowcolor{TableHeader}
\textbf{Exocentric source} & DINO$\uparrow$ & CLIP$\uparrow$ & FVD$\downarrow$ & Flow MSE$\downarrow$ \\
\midrule
No exo (Ego + Text)          & 58.73 & 38.72 & 204.62 & 3.19 \\
Exo at clip start            & 60.40 & 38.90 & 189.30 & 2.94 \\
\rowcolor{TableHighlight}Exo at clip end (main tables) & \textbf{61.25} & \textbf{39.30} & \textbf{182.25} & \textbf{2.83} \\
Random exo from another scene & 57.20 & 37.89 & 237.44 & 4.13 \\
\bottomrule
\end{tabular}
\end{adjustbox}
\end{table}

\subsection{Component Ablation with a Clip-Start Reference}
\label{app:clipstart}
To evaluate the contribution of each component without providing
future exocentric frames at inference, we repeat the component
ablation using an exocentric reference from the start of each clip.
All variants are evaluated on the same test examples with identical
random seeds and generation settings.
Table~\ref{tab:clipstart_ablation} shows that adding geometry weak
supervision to denoising fine-tuning improves all seven metrics,
and \graft{} provides further gains across every metric.
Relative to FT + GWS, \graft{} increases DINO-Score by 2.33
and CLIP-Score by 1.25, while reducing FVD by 36.47 and
Flow MSE by 1.09. Visual fidelity also improves, with SSIM
and PSNR increasing by 0.03 and 1.21, respectively, and LPIPS
decreasing by 0.03.
These results show that the benefits of geometry supervision
and \graft{} persist when the exocentric reference depicts only
the initial scene rather than the final state.\looseness-1

\begin{table}[t!]
\centering
\small
\setlength{\tabcolsep}{6pt}
\renewcommand{\arraystretch}{1.15}
\caption{\textbf{Component ablation with a clip-start exocentric reference.}
FT denotes denoising fine-tuning, and GWS denotes geometry weak
supervision. All variants receive the same conditioning inputs,
including an exocentric image from the start of the clip, and use
identical test examples, random seeds, and generation settings.
Adding GWS and then \graft{} improves all seven metrics without
access to future exocentric frames at inference.}
\label{tab:clipstart_ablation}
\begin{adjustbox}{max width=\linewidth}
\begin{tabular}{@{}lccccccc@{}}
\toprule
\rowcolor{TableHeader}
\textbf{Variant} & DINO$\uparrow$ & CLIP$\uparrow$ & SSIM$\uparrow$ & LPIPS$\downarrow$
& FVD$\downarrow$ & Flow MSE$\downarrow$ & PSNR$\uparrow$ \\
\midrule
FT only        & 55.96 & 36.70 & 0.73 & 0.22 & 267.94 & 4.93 & 21.67 \\
FT + GWS       & 58.07 & 37.65 & 0.75 & 0.19 & 225.77 & 4.03 & 22.62 \\
\rowcolor{TableHighlight}FT + GWS + \graft{}
& \textbf{60.40} & \textbf{38.90} & \textbf{0.78} & \textbf{0.16}
& \textbf{189.30} & \textbf{2.94} & \textbf{23.83} \\
\bottomrule
\end{tabular}
\end{adjustbox}
\end{table}

\subsection{Reward Component Ablation}
\label{app:reward_ablation}
We ablate each reward term in \graft{} while retaining the remaining terms. Table~\ref{tab:reward_ablation} reports the results. The full reward composition achieves the best performance across all seven metrics.
Removing $\mathcal{R}_{\text{per}}$ causes the largest deterioration in the visual-quality metrics, reducing SSIM from 0.79 to 0.77 and PSNR from 24.08 to 23.17, while increasing LPIPS from 0.15 to 0.18. Removing $\mathcal{R}_{\text{temp}}$ causes the largest decrease in temporal fidelity, increasing FVD from 182.25 to 213.25 and flow MSE from 2.83 to 3.70. Removing $\mathcal{R}_{\text{goal}}$ causes the largest decreases in semantic alignment, lowering DINO-Score from 61.25 to 59.62 and CLIP-Score from 39.30 to 38.49. Omitting $\mathcal{R}_{\text{env}}$ also degrades all seven metrics, consistent with the value of preserving scene context. These results support the complementary roles of the four reward terms. Appendix~\ref{app:reward_reliability} separately examines reward-judge agreement, agreement with human raters, robustness to the choice of judge, and a shortcut probe.\looseness-1

\begin{table*}[t!]
    \setlength{\tabcolsep}{8pt}
    \centering
    \caption{\textbf{Effect of reward components in \graft{}.} Each reward term contributes to improved semantic alignment and temporal coherence, while the full reward composition yields the strongest performance across all metrics.}
    \label{tab:reward_ablation}
    \resizebox{\textwidth}{!}{%
    \begin{tabular}{llccccccc}
        \toprule
\rowcolor{TableHeader}
        \textbf{Rewards} & \textbf{DINO-Score}\(\uparrow\) & \textbf{CLIP-Score}\(\uparrow\) & \textbf{SSIM} \(\uparrow\)  & \textbf{LPIPS} \(\downarrow\) & \textbf{FVD} \(\downarrow\) & \textbf{flow MSE} \(\downarrow\) & \textbf{PSNR} \(\uparrow\) \\
        \midrule
   \xmark$~~\mathcal{R}_{\text{goal}}$    & 59.62 & 38.49 & 0.78 & 0.16 & 205.96 & 3.48 & 23.48 \\
   \xmark$~~\mathcal{R}_{\text{env}}$     & 60.67 & 39.05 & 0.78 & 0.16 & 200.49 & 3.43 & 23.60 \\
    \xmark$~~\mathcal{R}_{\text{temp}}$    & 60.78 & 39.11 & 0.78 & 0.16 & 213.25 & 3.70 & 23.72 \\
    \xmark$~~\mathcal{R}_{\text{per}}$ & 60.32 & 38.80 & 0.77 & 0.18 & 204.13 & 3.48 & 23.17  \\
         \rowcolor{TableHighlight} \textbf{\modelname{} (Ours)} & \textbf{61.25} &  \textbf{39.30} &  \textbf{0.79}  &  \textbf{0.15} &  \textbf{182.25} &  \textbf{2.83} &  \textbf{24.08} \\     
        \bottomrule
    \end{tabular}
    }
\end{table*}

\section{Refinement across Video Backbones}
\label{app:matched}
We evaluate whether the benefits of \graft{} extend beyond
the Wan2.2 backbone used in \modelnamenc{}.
We equip Cosmos and HunyuanVideo with our structured conditioning
mechanism and geometry weak supervision, using the same inputs:
an initial egocentric frame, a goal instruction, and an auxiliary
exocentric image.
Table~\ref{tab:matched_backbones} compares each backbone before
and after \graft{} refinement, with the conditioning unchanged.
\graft{} reduces FVD by 15.9\% on Cosmos, 16.1\% on HunyuanVideo,
and 16.7\% on Wan2.2, while improving all six remaining metrics
on each backbone.
These consistent gains show that the refinement method benefits
all three evaluated backbones.
The Wan2.2-based \modelnamenc{} achieves the best results
across all seven metrics.
Wan2.2+CI is our Stage-1 model, so its comparison with
\modelnamenc{} measures the additional contribution of \graft{}.
This differs from the comparison with off-the-shelf Wan2.2, which measures the combined benefit of conditioning, geometry
supervision, and refinement.\looseness-1

\begin{table}[t!]
\centering
\small
\setlength{\tabcolsep}{6pt}
\renewcommand{\arraystretch}{1.15}
\caption{\textbf{Effect of \graft{} across video backbones.}
All three backbones use our structured conditioning mechanism
and geometry weak supervision (+CI), with the same egocentric
frame, goal instruction, and exocentric image as inputs.
Each backbone is evaluated before and after \graft{} refinement,
which improves all seven metrics.
Wan2.2+CI is our Stage-1 model; adding \graft{} yields the
full \modelnamenc{} model.}
\label{tab:matched_backbones}
\begin{adjustbox}{max width=\linewidth}
\begin{tabular}{@{}lccccccc@{}}
\toprule
\rowcolor{TableHeader}
\textbf{Model} & DINO$\uparrow$ & CLIP$\uparrow$ & SSIM$\uparrow$ & LPIPS$\downarrow$
& FVD$\downarrow$ & Flow MSE$\downarrow$ & PSNR$\uparrow$ \\
\midrule
Cosmos +CI              & 54.65 & 32.13 & 0.74 & 0.21 & 338.50 & 4.91 & 20.48 \\
Cosmos +CI +GRAFT         & 56.77 & 33.21 & 0.76 & 0.18 & 284.66 & 3.62 & 21.59 \\
\midrule
HunyuanVideo +CI        & 57.44 & 31.64 & 0.75 & 0.20 & 276.36 & 4.42 & 21.18 \\
HunyuanVideo +CI +GRAFT   & 59.37 & 32.75 & 0.77 & 0.17 & 231.89 & 3.28 & 22.20 \\
\midrule
Wan2.2 +CI (Stage-1)    & 58.92 & 38.05 & 0.76 & 0.18 & 218.72 & 3.92 & 22.87 \\
\rowcolor{TableHighlight}Wan2.2 +CI +GRAFT (\modelnamenc{})
& \textbf{61.25} & \textbf{39.30} & \textbf{0.79} & \textbf{0.15}
& \textbf{182.25} & \textbf{2.83} & \textbf{24.08} \\
\bottomrule
\end{tabular}
\end{adjustbox}
\end{table}

\section{Reliability of the VLM Reward Judge}
\label{app:reward_reliability}
We evaluate the consistency of the VLM reward scores used by
\graft{} by comparing GPT-5.5, our training-time judge, with
Gemini 3.1 Pro and Qwen3-VL 235B.
All three judges score the same 300 held-out generated videos,
comprising six rollouts for each of 50 conditions, using identical
prompts and sampled frames.
Table~\ref{tab:reward_cross_vlm} reports pairwise correlations
between their scores. GPT-5.5 has Spearman correlations of
0.83 with Gemini 3.1 Pro and 0.78 with Qwen3-VL 235B,
indicating that the training judge and both alternatives tend
to rank the evaluated rollouts similarly.
Table~\ref{tab:reward_icc} additionally reports intraclass
correlations and agreement on the highest-reward rollout
within each condition.\looseness-1

\subsection{Agreement across judges}
Table~\ref{tab:reward_icc} reports an ICC(2,1) of 0.60
for a single judge and an ICC(2,$k$) of 0.82 for the mean
of all three judges ($k=3$). The single-judge measure corresponds
to our training setup, which uses GPT-5.5 rather than an ensemble.
Before computing the refinement objective, \graft{} normalizes
rewards within each condition (Eq.~\eqref{eq:optimality}).
The training signal therefore reflects how each rollout scores
relative to the other candidates for the same condition,
rather than its absolute score.
Table~\ref{tab:reward_icc} also reports top-1 agreement,
which measures whether the judges select the same
highest-reward rollout within each condition.

\subsection{Robustness to the choice of judge}
We repeat \graft{} refinement using Qwen3-VL as the reward
judge in place of GPT-5.5. The resulting model achieves
an FVD of 187.9 and a DINO-Score of 60.8, compared with
182.25 and 61.25, respectively, when using GPT-5.5.
Although performance is slightly lower with Qwen3-VL,
both variants improve over Stage-1, which achieves an
FVD of 218.72 and a DINO-Score of 58.92.
Thus, \graft{} retains its advantage over Stage-1 when
the training judge is replaced.

\subsection{Agreement with human raters}
We compare the reward scores with the independently collected
ratings from the blind human study in Appendix~\ref{app:user}.
Table~\ref{tab:reward_human} pairs each reward component
with the corresponding human rating dimension.
Spearman correlations range from 0.71 to 0.79 across the
four components. The total reward also correlates with the
overall mean human rating ($\rho=0.80$, $\tau=0.66$),
indicating that higher-reward videos tend to receive
higher ratings from human evaluators.

\begin{table}[t!]
\centering
\small
\setlength{\tabcolsep}{10pt}
\renewcommand{\arraystretch}{1.15}
\caption{\textbf{Correlation between VLM reward scores.}
We report Spearman rank correlation ($\rho$) and Pearson
correlation ($r$) between each pair of judges on 300 held-out
generated videos (50 conditions, six rollouts per condition).
All judges evaluate the same sampled frames using identical
prompts. GPT-5.5 is the judge used during \graft{} refinement.}
\label{tab:reward_cross_vlm}
\begin{adjustbox}{max width=\linewidth}
\begin{tabular}{@{}lcc@{}}
\toprule
\rowcolor{TableHeader}
\textbf{Judge pair} & \textbf{Spearman $\rho$} & \textbf{Pearson $r$} \\
\midrule
GPT-5.5 and Gemini 3.1 Pro        & 0.83 & 0.81 \\
GPT-5.5 and Qwen3-VL 235B         & 0.78 & 0.76 \\
Gemini 3.1 Pro and Qwen3-VL 235B  & 0.76 & 0.74 \\
\bottomrule
\end{tabular}
\end{adjustbox}
\end{table}

\begin{table}[t!]
\centering
\small
\setlength{\tabcolsep}{10pt}
\renewcommand{\arraystretch}{1.15}
\caption{\textbf{Reliability of VLM reward scores across judges.}
ICC(2,1) measures reliability for a single judge, corresponding
to our training setup; ICC(2,$k$), with $k=3$, measures
reliability when averaging scores from all three judges.
Top-1 agreement measures how often all three judges select
the same highest-reward rollout among the six candidates
for a condition.}
\label{tab:reward_icc}
\begin{adjustbox}{max width=\linewidth}
\begin{tabular}{@{}lc@{}}
\toprule
\rowcolor{TableHeader}
\textbf{Reliability measure} & \textbf{Value} \\
\midrule
ICC(2,1), a single judge, the configuration used for training & 0.60 \\
ICC(2,k), mean of the three judges & 0.82 \\
Within-condition top-1 agreement, $K=6$ rollouts & 81.4\% \\
\bottomrule
\end{tabular}
\end{adjustbox}
\end{table}

\begin{table}[t!]
\centering
\small
\setlength{\tabcolsep}{10pt}
\renewcommand{\arraystretch}{1.15}
\caption{\textbf{Correlation between reward scores and human ratings.}
We report Spearman rank correlation ($\rho$) and Kendall
rank correlation ($\tau$) between each reward component
and the corresponding rating dimension from the blind human
study in Appendix~\ref{app:user}.
The final row compares the total reward $R_{\mathrm{total}}$
with the overall mean human rating.}
\label{tab:reward_human}
\begin{adjustbox}{max width=\linewidth}
\begin{tabular}{@{}llcc@{}}
\toprule
\rowcolor{TableHeader}
\textbf{Reward term} & \textbf{Human dimension} & \textbf{Spearman $\rho$} & \textbf{Kendall $\tau$} \\
\midrule
$\mathcal{R}_{\text{goal}}$ & Alignment          & 0.79 & 0.64 \\
$\mathcal{R}_{\text{env}}$  & Smooth Environment & 0.73 & 0.58 \\
$\mathcal{R}_{\text{temp}}$ & Smooth Motion      & 0.71 & 0.56 \\
$\mathcal{R}_{\text{per}}$  & Fidelity           & 0.76 & 0.61 \\
\midrule
$\mathcal{R}_{\text{total}}$ & Overall mean      & 0.80 & 0.66 \\
\bottomrule
\end{tabular}
\end{adjustbox}
\vspace{-0.3cm}
\end{table}

\subsection{Reward hacking}
We first examine whether \graft{} improves evaluation measures
that are not used to compute the training reward.
Under the clip-start evaluation in
Table~\ref{tab:clipstart_ablation}, adding \graft{} to FT + GWS
increases DINO-Score from 58.07 to 60.40 and CLIP-Score
from 37.65 to 38.90, reduces Flow MSE by 27.0\%
(from 4.03 to 2.94), and increases SSIM from 0.75 to 0.78.
None of these four metrics contributes to the reward.
The independently collected blind human ratings in
Table~\ref{tab:user_study_stage1} also improve after refinement.
Together, these results provide evidence that the benefits of
\graft{} extend beyond the scores it directly optimizes.
We do not use improvements in PSNR, FVD, or LPIPS as
additional evidence here, since these metrics contribute
to $\mathcal{R}_{\text{per}}$.

We also test whether the reward favors a shortcut that depicts
the target state without performing the required actions.
We construct a static video by repeating the target-state image
throughout the clip. Although this video receives a visual-alignment
score of 0.82 within $\mathcal{R}_{\text{goal}}$, it scores only
0.21 on temporal causality ($\mathcal{R}_{\text{temp}}$) and
0.30 on environment consistency ($\mathcal{R}_{\text{env}}$).
Its composite reward is 0.44, below the median of 0.68 for
generated rollouts. Thus, in this test, matching the target image
without generating the required action sequence does not yield
a high overall reward.

\begin{figure}[t!]
\centering
\includegraphics[width=0.93\linewidth]{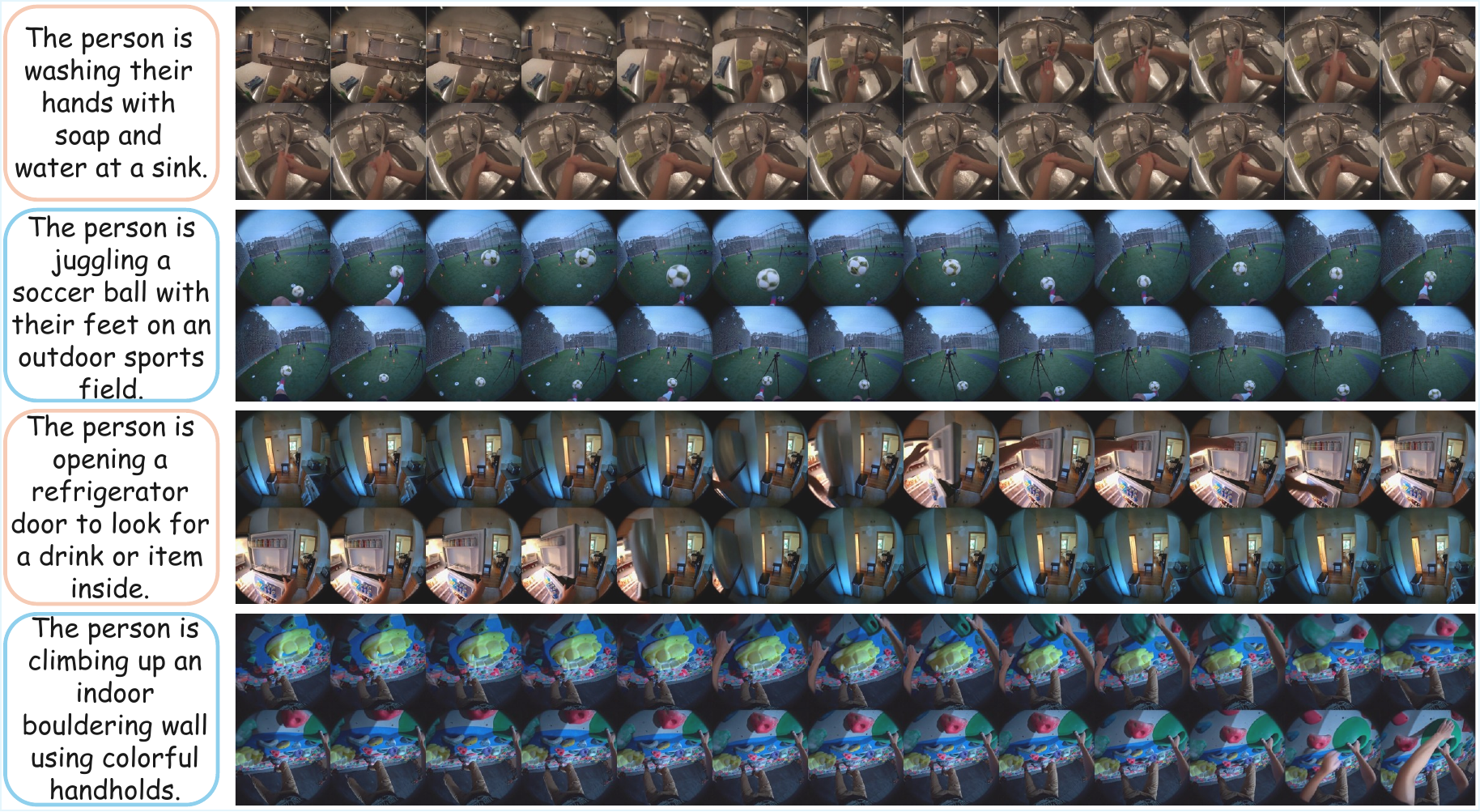} 
\caption{\textbf{Qualitative egocentric video rollouts.} 
\modelnamenc{} generates temporally coherent first-person video trajectories that follow the intended activity while preserving scene structure and realistic hand–object interactions across diverse environments.}
\label{fig:qualfig}
\vspace{-0.2cm}
\end{figure}

\begin{figure}[!t]
\centering
    \includegraphics[width=0.93\textwidth]{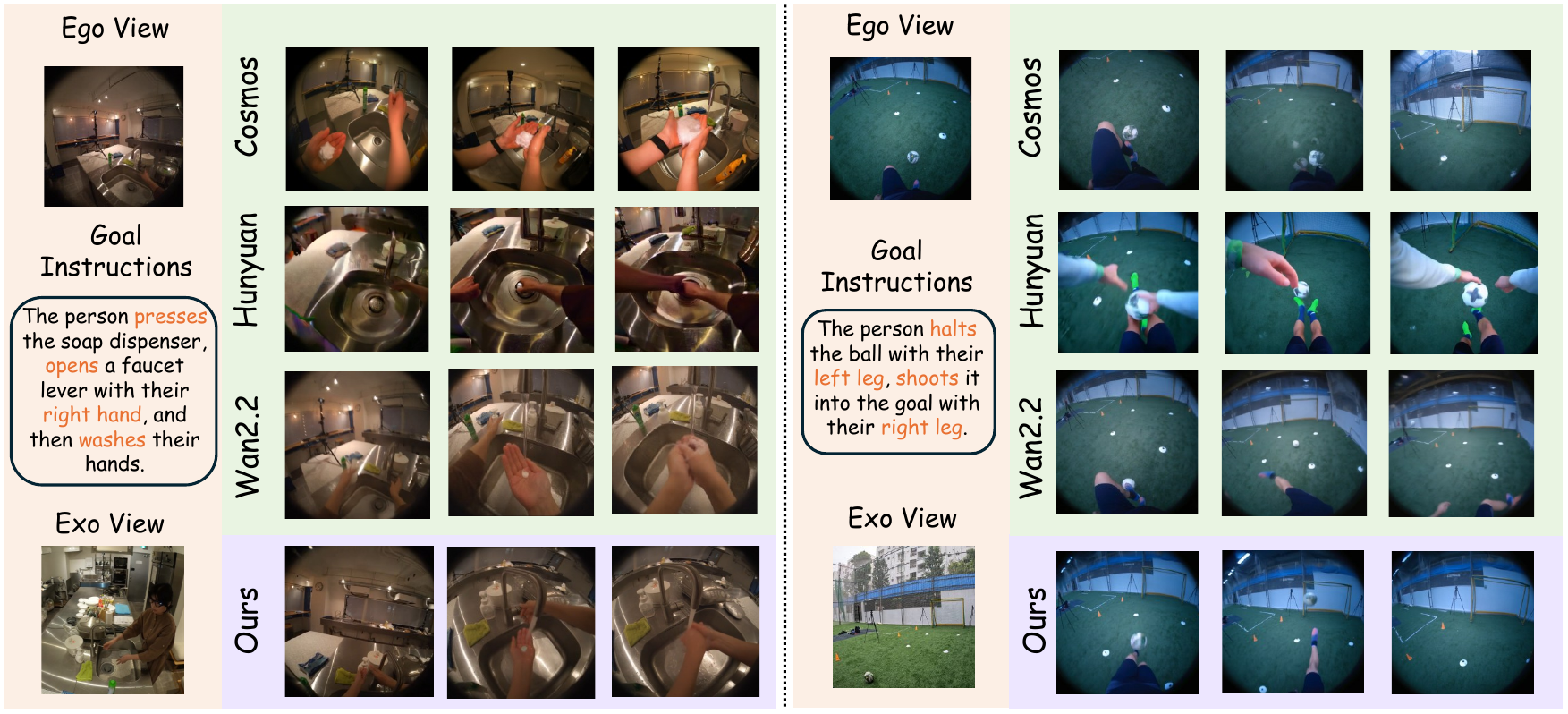}
\captionof{figure}{\textbf{Qualitative Comparison between \modelname{} and baselines}. Sample frames from generated videos illustrating two scenarios. Top row: In the hand-washing task, baselines struggle with object consistency (\eg Cosmos hallucinates the soap source) or ignore scene context (\eg Wan2.2 bbypasses the soap on the table), while EgoForge successfully executes the action. using existing objects. Bottom row: In the soccer task, baselines exhibit severe artifacts like ghosting (Cosmos) or fail to follow precise instructions regarding motion and goals (Hunyuan, Wan2.2). \modelnamenc{} accurately executes the complex command: trapping with the left leg and shooting with the right.}
\label{fig:quali2}
\vspace{-0.2cm}
\end{figure}

\begin{figure}[!t]
    \centering
\includegraphics[width=0.99\linewidth,keepaspectratio]{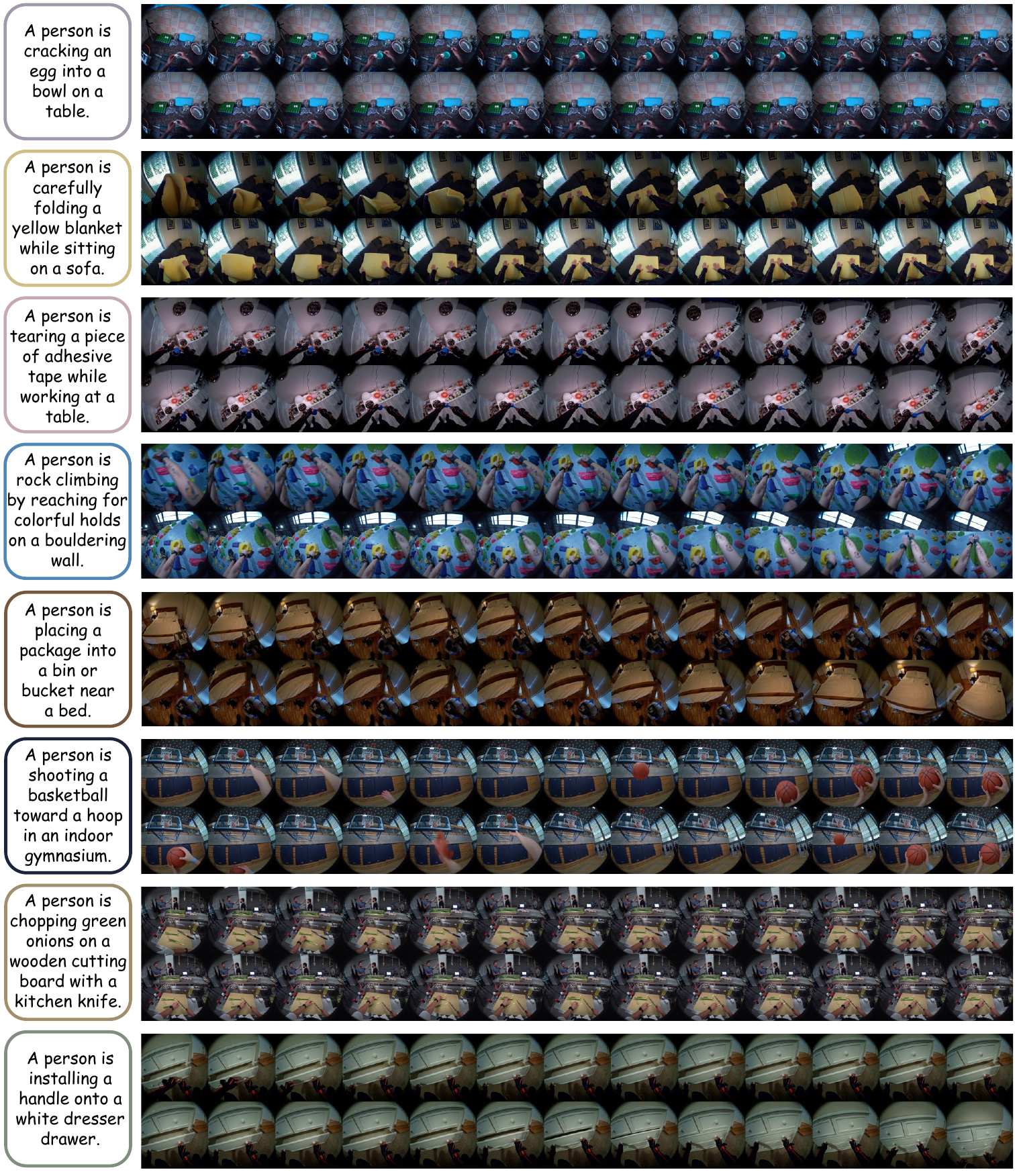}
    \caption{\textbf{Dense visualization of long-duration sequences.}
    For each \modelnamenc{}-generated video, we show 26 uniformly sampled frames to illustrate how actions unfold over time. Across diverse first-person tasks, \modelnamenc{} preserves scene structure, maintains a stable egocentric viewpoint, and produces smooth, goal-consistent hand--object interactions.}
    \label{fig:long_sequence_appendix}
\end{figure}

\section{Additional Qualitative Examples}
\label{sec:vis_results}
Across diverse activities and environments, the generated rollouts follow the specified instructions while maintaining stable scene structure and consistent egocentric viewpoint motion. Importantly, complex multi-step behaviors (\eg manipulating objects, interacting with appliances, or performing physical activities) unfold naturally over time, demonstrating that \modelnamenc{} captures both the procedural structure of actions and the surrounding scene context required for goal-directed egocentric simulation. 
In more dynamic scenarios like rock climbing or shooting a basketball, the rollouts exhibit clear temporal progression toward visually identifiable goals while maintaining consistent environment geometry, including climbing holds and the basketball hoop. These examples highlight the ability of \modelnamenc{} to generate temporally coherent, goal-directed egocentric sequences that sustain interaction dynamics and scene consistency over long durations.

\section{Device for Real World Experiments}
\label{sec:devices}
\begin{wrapfigure}{r}{0.6\columnwidth}
    \centering
    \includegraphics[width=\linewidth]{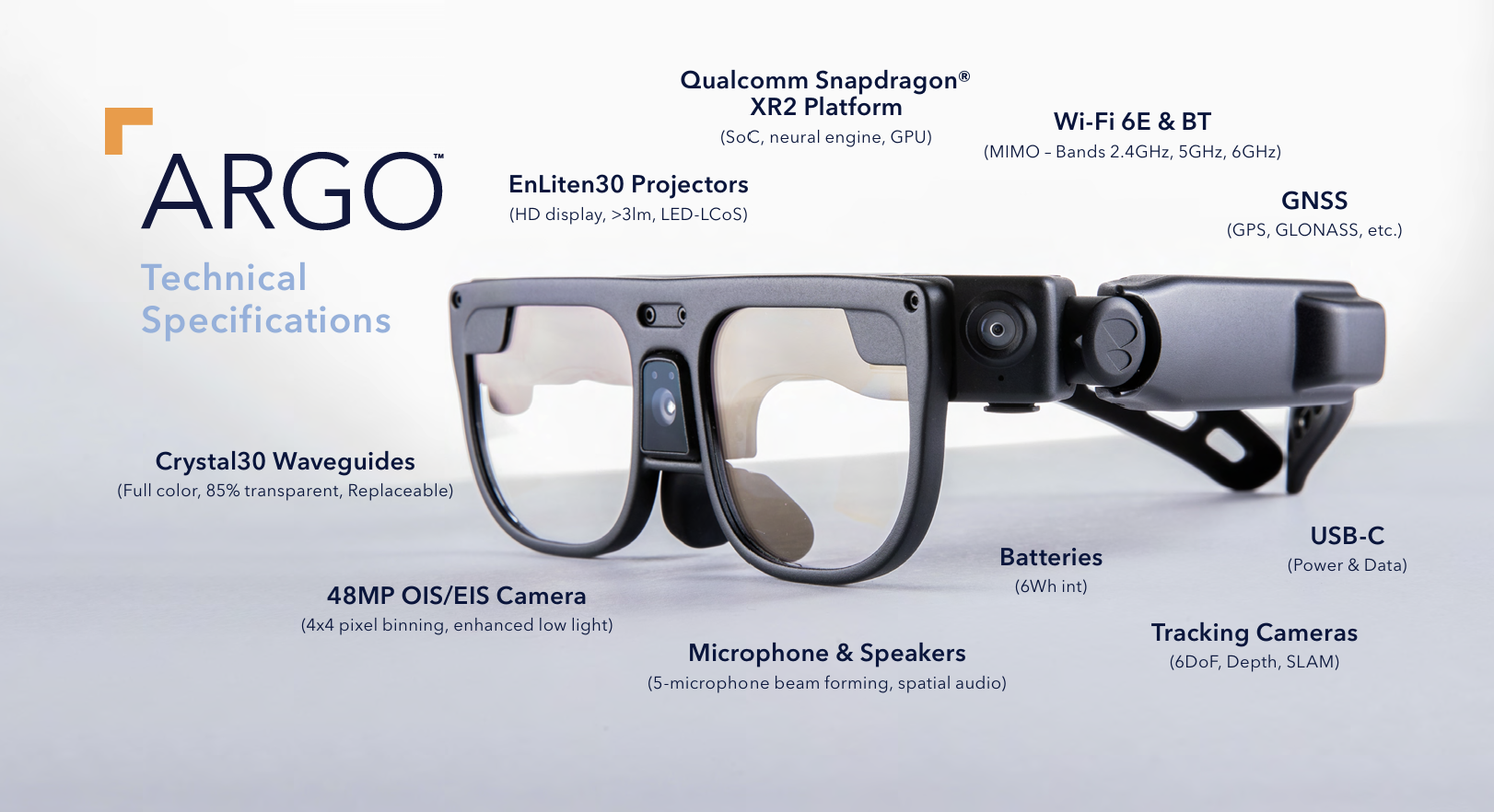}
    \caption{\textbf{DigiLens ARGO Smart glasses}}
    \label{fig:glass}
\end{wrapfigure}For our real-world experiments, we used the DigiLens ARGO smartglasses shown in Figure \ref{fig:glass} to capture egocentric video data. The ARGO smartglasses employ Augmented Reality (AR) technology, which creates a visual overlay on the real world seen through the transparent display. This device is an enterprise-grade standalone system based on DigiLens’ core waveguide technology and is designed for industrial and enterprise use cases. The ARGO provides several key technical advantages for high-quality data capture, including a 48MP camera with autofocus, optical image stabilization (OIS), and electronic image stabilization (EIS), supporting $4\times4$ pixel binning and enhanced low-light performance. It also features a sophisticated five-microphone beamforming array designed to pick up the wearer's voice in noisy environments and provides integrated stereo spatial audio recordings. The device runs on the Qualcomm Snapdragon XR2 Platform, providing onboard mobile computing capabilities. The ARGO smartglasses are critical to our evaluation because they allow us to test EgoForge’s robustness in real-world, out-of-domain (OOD) settings, addressing a limitation of previous egocentric world models. The high-fidelity data captured by the ARGO's advanced sensor suite provides the essential grounding needed for \modelnamenc{} to reliably transfer visual cues and follow high-level semantic intent, producing coherent and controllable rollouts despite significant real-world variability.\looseness-1

\section{Broader Impacts}\label{app:impacts}
\modelnamenc{} aims to advance egocentric world simulation from sparse wearable observations, enabling controllable first-person video rollouts that can support future research in XR, assistive AI, human–AI collaboration, and embodied learning. Potential positive impacts include safer and lower-cost prototyping of interactive AR/VR systems, synthetic data generation for egocentric perception, and simulation-based evaluation of task-oriented agents without requiring repeated real-world data collection. At the same time, realistic egocentric video generation may introduce risks if misused, such as creating misleading first-person footage, simulating private environments, or enabling surveillance-like applications when combined with wearable sensors. The use of egocentric data also raises privacy considerations, since first-person views may contain bystanders, personal spaces, or sensitive objects. We therefore view \modelnamenc{} as a research tool for simulation and evaluation rather than for identification, surveillance, or decision-making about individuals. Responsible deployment should include consented and licensed data sources, privacy filtering or de-identification when appropriate, disclosure or watermarking of synthetic videos, restrictions against impersonation or surveillance use cases, and human oversight in safety-critical applications.

 \clearpage
\section{Evaluation and Annotation Prompts}\label{app:prompts}
\begin{planbox}{Prompt for \graft{} Reward}
\small
You are an expert evaluator assessing goal achievement in egocentric video generation. Your task is to evaluate whether a generated first-person video successfully accomplishes a specified manipulation task by analyzing three dimensions: Goal Alignment, Environment Preservation, and Temporal Causality. You will evaluate a process video alongside its initial state image, target state image, and task description, providing quantitative scores for each dimension based on the criteria specified below. \\

\promptheading{The first task: Goal Alignment Scoring (Total 0–2.0 points)} \\
   Assess goal achievement through task completion analysis and semantic alignment between the video final state and the target state:
      \begin{itemize}[itemsep=0em, leftmargin=3.5em]
          \item \textbf{Task Completion Score:} Assign scores from 0.0 to 1.0 based on state transition analysis from initial to final frames, where 0.7$\sim$1.0 indicates all or most required state changes achieved, 0.4$\sim$0.6 indicates partial achievement with some changes incomplete, and 0.0$\sim$0.3 indicates minimal or no required changes with task not accomplished.
          \item \textbf{Visual Alignment Score:} Assign scores from 0.0 to 1.0 based on semantic correspondence between video final state and target image (accounting for first-person vs third-person viewpoint difference), where 0.7$\sim$1.0 indicates all or most target elements and spatial relationships present, 0.4$\sim$0.6 indicates some elements present with others missing, and 0.0$\sim$0.3 indicates minimal or no target elements present.
      \end{itemize}
   Provide a detailed breakdown for task completion and visual alignment. \\
   You need to give the score with the following format: \\
   \texttt{{\{"score": [task completion score, visual alignment score]\}}}\\

\promptheading{The second task: Environment Preservation Scoring (Total 0–2.0 points)} \\
  Assess environment consistency with initial state and resistance to target image contamination:
      \begin{itemize} [itemsep=0em, leftmargin=3.5em]
          \item \textbf{Consistency Score:} Assign scores from 0.0 to 1.0 based on environment consistency between initial and video states (ignoring target image environment which may differ), where 0.7$\sim$1.0 indicates all or most environmental features unchanged, 0.4$\sim$0.6 indicates some features consistent with moderate drift, and 0.0$\sim$0.3 indicates most features changed with major inconsistencies.
          \item \textbf{Contamination Score:} Assign scores from 0.0 to 1.0 based on freedom from target image environmental contamination, where 0.7$\sim$1.0 indicates no or minimal contamination with video maintaining original environment, 0.4$\sim$0.6 indicates moderate contamination with some target features leaked, and 0.0$\sim$0.3 indicates heavy contamination with video heavily influenced by target's different environment.
      \end{itemize}
   Provide a detailed breakdown for environmental consistency assessment. \\
   You need to give the score with the following format: \\
   \texttt{{\{"score": [consistency score, contamination score]\}}}\\

\promptheading{The third task: Temporal Causality Scoring (Total 0–2.0 points)} \\
   Assess whether the video demonstrates physically plausible and causally coherent progression from initial to final state:
      \begin{itemize}[itemsep=0em, leftmargin=3.5em]
          \item \textbf{Physics Plausibility Score:} Assign scores from 0.0 to 1.0 based on adherence to real-world physics, where 0.7$\sim$1.0 indicates all or most actions are physically realistic with proper gravity and motion, 0.4$\sim$0.6 indicates some actions are realistic with several implausibilities, and 0.0$\sim$0.3 indicates most actions violate physics with impossible movements.
          \item \textbf{Causal Logic Score:} Assign scores from 0.0 to 1.0 based on clarity of cause-effect relationships, where 0.7$\sim$1.0 indicates all or most actions have clear visible causes with logical sequences, 0.4$\sim$0.6 indicates some actions have visible causes with a partial causal chain, and 0.0$\sim$0.3 indicates most actions lack visible causes with broken logic.
      \end{itemize}
   Provide a detailed breakdown for each dimension. \\
   You need to give the score with the following format: \\
   \texttt{{\{"score": [physics plausibility score, causal logic score]\}}}\\

\end{planbox}

\begin{planbox}[EgoPeriwinkle]{Caption Refinement Prompt. The model takes both the original coarse caption and the video frames as input to generate a visually grounded fine-grained description.}
\small
You are an expert visual data annotator for ego-centric video datasets. Your task is to rewrite and refine a video text annotation by synthesizing the provided \textbf{Original Caption} with the actual \textbf{Video Content}.

\textbf{Your Output Goal}: A vivid, photorealistic description (3--4 sentences) that strictly aligns with the visual evidence in the video while maintaining the semantics of the original caption.

\textbf{Instructions:}
\begin{enumerate}
    \item \textbf{Visual Grounding}: Analyze the \texttt{Video Input} first. Correct the \texttt{Original Caption} if it contradicts the visual evidence (e.g., wrong object color, wrong hand usage).
    \item \textbf{Structure}:
    \begin{itemize}
        \item \textbf{Sentence 1 (Setup)}: Describe the visible appearance of the hands and objects (e.g., material, texture) based on the video start frame.
        \item \textbf{Sentence 2 (Action)}: Describe the fine-grained motion trajectory observed in the video clips.
        \item \textbf{Sentence 3 (Reaction)}: Describe the object's physical response (deformation, displacement) shown in the video.
        \item \textbf{Sentence 4 (Outcome)}: Conclude with the final state visible in the end frame.
    \end{itemize}
    \item \textbf{Constraints}: Do not hallucinate details not present in the video.
\end{enumerate}

\textbf{Example Input}:
\begin{itemize}
    \item \textbf{Original Caption}: ``The person pours water.''
    \item \textbf{Video Input}: [Sequence of frames showing a hand tilting a blue cup]
\end{itemize}

\textbf{Example Output}: ``The person's hand, gripping a ribbed blue plastic cup, tilts it sharply over a white bowl. A clear liquid cascades from the cup... [omitted for brevity] ...leaving droplets on the rim.''

\hrulefill

\textbf{Input Data}:
\begin{description}
    \item[\textbf{Original Caption}:] \texttt{\{input\_caption\}}
    \item[\textbf{Video Input}:] \texttt{<video\_frames>}
\end{description}

\end{planbox}

\begin{planbox}[EgoTerracotta]{Video Segmentation Prompt for action localization} 
\small

You are an expert temporal-action summarizer. You are given non-overlapping 10-second segments derived from a long video. Each segment includes: 

      \begin{itemize}[itemsep=0em, leftmargin=3.5em]
          \item start and end: segment time in seconds (absolute timestamps from the video).
          \item raw: concatenated human annotations overlapping that 10-second window.
          \item filtered: the same annotations with any sentences containing disallowed categories removed, to highlight stationary/hand-object actions.
      \end{itemize}
Your task:

\begin{enumerate}
    \item Select 3 to 4 distinct stationary atomic actions across the timeline.
    
    \item ``Stationary'' means the action is performed in place (hand/object manipulations are preferred).
    
    \item The set of allowed stationary actions is NOT fixed. Infer candidate actions from the provided text (e.g., \textit{shuffle, deal, draw, put, take, hold, open, close, align, sort, count, stack, fan, flip/turn over, organize, adjust, press, push, pull, slide, rotate an object, etc.}).
    
    \item The following categories are strictly disallowed for selection as the main action: \texttt{\{disallowed\_list\}}.
    
    \item Avoid describing locomotion (walking/moving/stepping/running), speech (talking/speaking/chatting), or waiting/idle states.
    
    \item Prefer segments where a single hand-on-object action clearly dominates the 10-second window. Use the ``\texttt{filtered}'' field primarily to judge this. If ``\texttt{filtered}'' is empty for a segment, skip it.
    
    \item Each chosen output must represent a different action type (no duplicates). If the same action appears in multiple segments, choose only one representative segment.
    
    \item For each selected segment, write exactly one concise English sentence that describes only the atomic action (single main verb phrase). Do not include walking, talking, or waiting. Avoid chaining multiple verbs. Keep it specific and objective (no speculation).
    
    \item IMPORTANT: Use ``the person'' as the subject in all descriptions. Do NOT use ``she'', ``he'', ``C'', or any other pronouns or names.
\end{enumerate}

Formatting requirements:
\begin{itemize}[itemsep=0em, leftmargin=3.5em]
    \item Output ONLY a JSON array (no other text).
    \item Each array element MUST be an object with keys: "start", "end", "text".
    \item "start" and "end" MUST copy the segment's start and end values (integer timestamps in seconds).
    \item "text" MUST be a single English sentence describing only the stationary atomic action.
    \item Return exactly 5 to 8 objects. If fewer than 5 valid stationary actions exist, return as many as valid (possibly an empty array).
\end{itemize}

Segments JSON:
\texttt{\{segments\_json\}}

\end{planbox}

\end{document}